\documentclass{article}

\usepackage[preprint]{neurips}
\usepackage[table,xcdraw]{xcolor}
\usepackage{tcolorbox}
\usepackage{booktabs}
\usepackage{tabularx}
\usepackage{colortbl}
\usepackage{caption}

\definecolor{crIndoor}{RGB}{255, 200, 196}
\definecolor{crNature}{RGB}{197, 225, 222}
\definecolor{crUrban}{RGB}{250, 234, 219}

\newtcbox{\crbx}[1]{on line, 
colback=#1, colframe=#1!60!black,
boxrule=0.2pt, arc=1pt, boxsep=0pt,
left=2pt,right=2pt,top=1pt,bottom=1pt}

\usepackage[utf8]{inputenc} 
\usepackage[T1]{fontenc}    
\usepackage{url}            
\usepackage{booktabs}       
\usepackage{amsfonts}       
\usepackage{nicefrac}       
\usepackage{microtype}      
\usepackage{xcolor}         
\usepackage{subcaption} 

\usepackage{graphicx}
\usepackage{booktabs}

\usepackage[accsupp]{axessibility}  

\usepackage{graphicx}

\usepackage{tikz}
\usepackage{comment}
\usepackage{amsmath,amssymb} 
\usepackage{bbm}
\usepackage{booktabs}
\usepackage[colorlinks]{hyperref}

\usepackage{algorithm}

\usepackage{mdframed}
\usepackage{mdframed}
\usepackage{amsmath, amssymb}

\usepackage{orcidlink}

\usepackage{amstext}
\usepackage{amssymb}
\usepackage{booktabs}

\usepackage{tikz}
\usepackage{comment}
\usepackage{amsmath,amssymb} 
\usepackage{bbm}
\usepackage{booktabs}
\usepackage{graphicx}
\usepackage{bm}
\usepackage{multirow}
\usepackage{cases}
\usepackage{color}
\usepackage{ulem}
\usepackage[table]{xcolor} 
\definecolor{dino}{RGB}{249,231,227}

\usepackage{url}
\usepackage{epsfig}
\usepackage{amsmath}
\usepackage{amssymb}
\usepackage{amsthm}
\usepackage{multirow}
\usepackage{soul}
\usepackage{footnote}
\usepackage{tablefootnote}
\usepackage{caption}
\usepackage{subcaption}
\usepackage{mathtools}

\usepackage{url}
\usepackage{multirow}
\usepackage{booktabs}
\usepackage{pifont}
\usepackage{xcolor}
\usepackage{graphicx}
\usepackage{amsmath}
\usepackage{algorithm}
\usepackage{mathtools}
\usepackage{algorithm}
\usepackage{algpseudocode}
\usepackage{listings}
\usepackage{wrapfig}
\usepackage{subcaption}
\usepackage{xcolor,colortbl}

\usepackage{marvosym} 

\definecolor{top1}{RGB}{255,179,179}
\definecolor{top2}{RGB}{255,217,179}
\definecolor{top3}{RGB}{255,255,179}

\usepackage{tikz}
\usepackage{comment}
\usepackage{amsmath,amssymb} 
\usepackage{bbm}
\usepackage{booktabs}
\usepackage[colorlinks]{hyperref}
\usepackage{graphicx}
\usepackage{bm}
\usepackage{multirow}

\usepackage{xcolor,pifont}
\newcommand*\colourcheck[1]{%
  \expandafter\newcommand\csname #1check\endcsname{\textcolor{#1}{\ding{52}}}%
}
\colourcheck{blue}
\colourcheck{green}
\colourcheck{red}
\usepackage{makecell}

\usepackage{xcolor}
\usepackage{graphicx}
\usepackage{caption}
\usepackage{booktabs}

\definecolor{crIndoor}{RGB}{255, 200, 196}
\definecolor{crNature}{RGB}{197, 225, 222}
\definecolor{crUrban}{RGB}{250, 234, 219}

\title{
WonderFree: Enhancing Novel View Quality and Cross-View Consistency for 3D Scene Exploration
}

\author{
~~~~~~Chaojun Ni\footnotemark[1]~\textsuperscript{\rm 1,2}
~~~~~~Jie Li\footnotemark[1]~\textsuperscript{\rm 1}
~~~~~~Haoyun Li\footnotemark[1]~\textsuperscript{\rm 1,3}
~~~~~~Hengyu Liu\footnotemark[1]~\textsuperscript{\rm 1,4} \\
~~~~~~ \textbf{Xiaofeng Wang}~\textsuperscript{\rm 1} 
~~~~~~ \textbf{Zheng Zhu}\footnotemark[2]~~\textsuperscript{\rm 1}
~~~~~~ \textbf{Guosheng Zhao}~\textsuperscript{\rm 1,3} 
~~~~~~ \textbf{Boyuan Wang}\textsuperscript{\rm 1,3} \\
~~~~~~ \textbf{Chenxin Li}\textsuperscript{\rm 1,4}
~~~~~~ \textbf{Guan Huang}\textsuperscript{\rm 1}
~~~~~~ \textbf{Wenjun Mei}\footnotemark[2]~~\textsuperscript{\rm 2}
\\
\textsuperscript{\rm 1}GigaAI
~ ~ \textsuperscript{\rm 2}Peking University \\
~ ~ \textsuperscript{\rm 3}Institute of Automation, Chinese Academy of Sciences \\
~ ~ \textsuperscript{\rm 4}The Chinese University of Hong Kong
\vspace{2mm}\\
Project Page: \textcolor[HTML]{FF1493}{\url{https://Wonder-Free.github.io/}}}

\begin{document}
\maketitle
\renewcommand{\thefootnote}{\fnsymbol{footnote}}
\footnotetext[1]{
These authors contributed equally to this work. 
}
\footnotetext[2]{\mbox{Corresponding authors. zhengzhu@ieee.org, mei@pku.edu.cn.}}

\begin{figure*}[ht]
    \centering
    \vspace{-1em}
    \includegraphics[width=.9\linewidth]{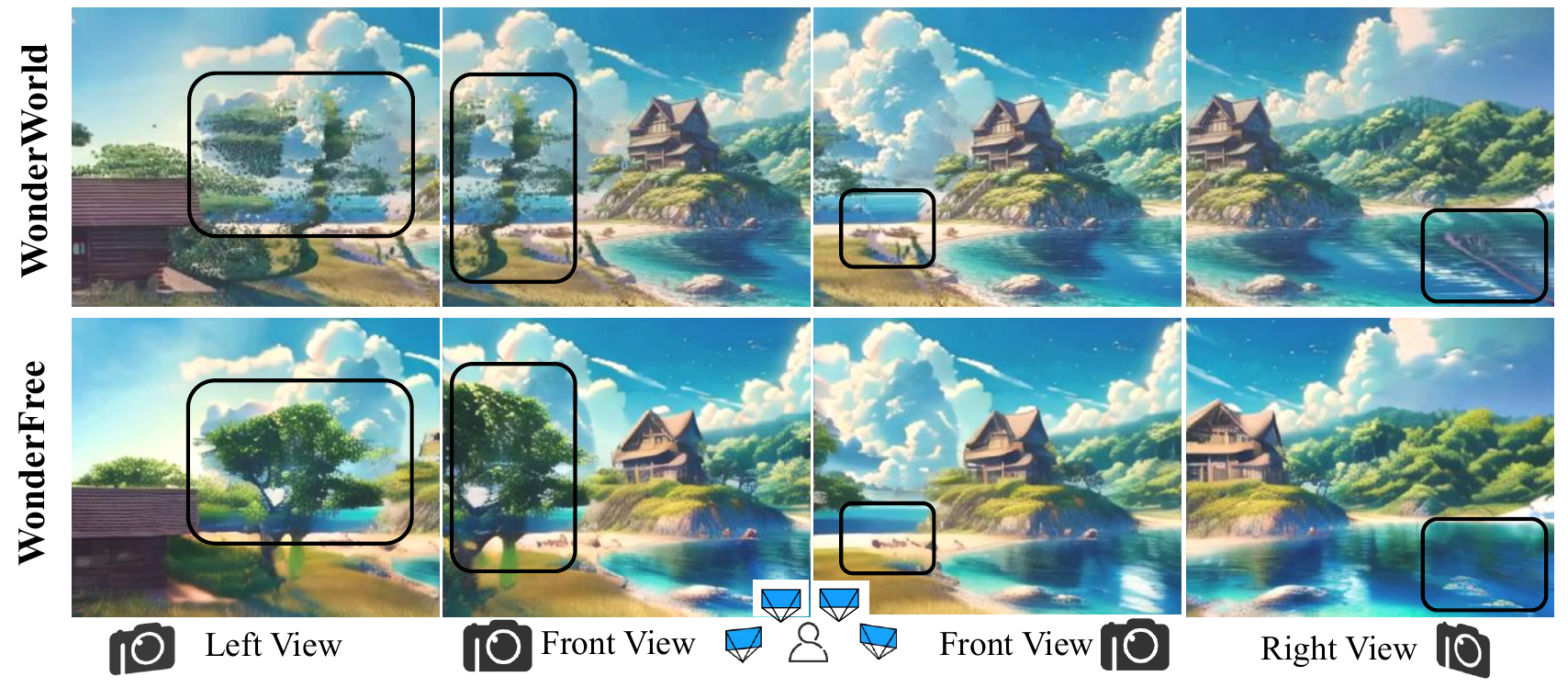} 
    \vspace{1em}
    \caption{We compare our method with WonderWorld~\cite{wonderworld} under novel views, including front and side perspectives. The black-box regions in WonderWorld~\cite{wonderworld} exhibit noticeable floaters and distortions, whereas WonderFree maintains clarity and visual consistency.}
\label{fig:teaser}
\end{figure*}


\begin{abstract}
Interactive 3D scene generation from a single image has gained significant attention due to its potential to create immersive virtual worlds. However, a key challenge in current 3D generation methods is the limited explorability, which cannot render high-quality images during larger maneuvers beyond the original viewpoint, particularly when attempting to move forward into unseen areas.
To address this challenge, we propose WonderFree, the first model that enables users to interactively generate 3D worlds with the freedom to explore from arbitrary angles and directions. Specifically, we decouple this challenge into two key subproblems: novel view quality, which addresses visual artifacts and floating issues in novel views, and cross-view consistency, which ensures spatial consistency across different viewpoints. To enhance rendering quality in novel views, we introduce WorldRestorer, a data-driven video restoration model designed to eliminate floaters and artifacts. In addition, a data collection pipeline is presented to automatically gather training data for WorldRestorer, ensuring it can handle scenes with varying styles needed for 3D scene generation. Furthermore, to improve cross-view consistency, we propose ConsistView, a multi-view joint restoration mechanism that simultaneously restores multiple perspectives while maintaining spatiotemporal coherence. Experimental results demonstrate that WonderFree not only enhances rendering quality across diverse viewpoints but also significantly improves global coherence and consistency. These improvements are confirmed by CLIP-based metrics and a user study showing a 77.20\% preference for WonderFree over WonderWorld, enabling a seamless and immersive 3D exploration experience. The code, model, and data will be publicly available.

\end{abstract}

\section{Introduction}
\label{submission}

Creating a 3D world from text descriptions or a single image and freely exploring it has long been a human dream. In recent years, many studies~\cite{wonderworld,text2room,luciddreamer,genex,pano2room,dreamscene360,diffpano,wonderturbo,diffusion360} have made significant progress in this field. However, current methods for generating 3D scenes still have obvious limitations in terms of explorability. These generated 3D worlds can often only render high-quality images from a few specific viewpoints, making it challenging for users to explore freely and continuously from any angle and direction, which severely limits their immersion and practicality.

The first major challenge for limited explorability lies in novel view synthesis~\cite{mvsplat,viewcrafter,layerview,layerdepth,dust3r,wondersingle,synsin}. When users freely explore within the generated 3D world, the system must ensure high-quality image generation from every perspective, particularly when the viewpoint moves forward beyond the original viewing range. However, existing 3D scene generation methods~\cite{text2room,luciddreamer} typically involve first generating multi-view images or panoramas of the scene, which are subsequently converted into 3D representations. These approaches use a limited number of views to supervise the generation of 3D worlds, making it difficult to maintain quality and consistency when confronted with significant viewpoint variations.

The second core challenge lies in ensuring multi-view consistency in the generated 3D world~\cite{worldscore,sculpt3d,gaussctrl,sv4d,viewfusion,Multiview,drivedreamer2}. For a continuous and interactive 3D environment, it is crucial to maintain high consistency across different perspectives in terms of geometric structure, semantic information, and visual appearance. However, existing methods~\cite{wonderworld,wonderturbo,wonderjourney} primarily focus on optimizing single-view output quality, often neglecting cross-view consistency constraints. These methods process each viewpoint independently, disregarding the spatial consistency and geometric relationships between different perspectives.

To address these challenges, we introduce WonderFree, the first 3D world generation model that enables users to freely explore from arbitrary angles and directions. Specifically, WonderFree first generates a coarse 3D world, and then progressively refines it through iterative optimization. In each iteration, videos from novel viewpoints are rendered, which may be filled with floaters and artifacts. These videos are then restored to obtain high-quality novel views, which are used to further refine the 3D world.  To eliminate ghosting artifacts and distortions in novel views of the generated 3D worlds, we propose WorldRestorer, a restoration model fine-tuned from a video generation framework~\cite{svd}. For effective training of WorldRestorer, we design an automated data generation pipeline that constructs video restoration datasets across a diverse range of scene styles. During fine-tuning, we introduce a masking strategy to enhance the restoration of occluded regions, thereby ensuring high-quality rendering even under large viewpoint maneuvers. Meanwhile, to ensure consistency across different viewpoints, we further integrate the ConsistView mechanism into WorldRestorer, a multi-view joint restoration approach. This mechanism aims to unify spatial coherence within individual views and across views, significantly improving the overall spatiotemporal consistency and immersion of the generated 3D worlds.

We present rendered images obtained by substantially translating and subsequently rotating the viewpoint. Specifically, the camera is first significantly shifted from its initial position, followed by a rotation to capture four distinct viewpoints, as shown in Fig.~\ref{fig:teaser}. Existing state-of-the-art interactive 3D generation methods~\cite{wonderworld,wonderturbo,wonderjourney}, such as WonderWorld~\cite{wonderworld}, fail to render high-quality novel views under these large viewpoint transformations, exhibiting noticeable floaters, artifacts, and structural distortions (e.g., deformation of trees). In contrast, WonderFree robustly handles extensive viewpoint translations and rotations, effectively preserving spatial consistency across multiple perspectives.


The main contributions of this paper are summarized as follows:

\begin{itemize}
    \item We introduce WonderFree, a framework enabling users to freely explore 3D worlds from arbitrary angles and directions. This is achieved by designing an automated data generation pipeline and proposing WorldRestorer, a restoration model fine-tuned from a video generation framework to eliminate ghosting artifacts and distortions.
    \item For maintaining multi-view consistency, we integrate the ConsistView mechanism into WorldRestorer, which enhances spatial coherence both within individual views and across different viewpoints through a multi-view joint restoration approach.
    \item We conduct comprehensive experiments to demonstrate that WonderFree not only enhances the quality of novel view synthesis but also ensures multi-view consistency, achieving state-of-the-art performance on CLIP-based metrics and user study preference rates.
    
\end{itemize}

\begin{figure*}[t]
    \centering
    \includegraphics[width=0.99\linewidth]{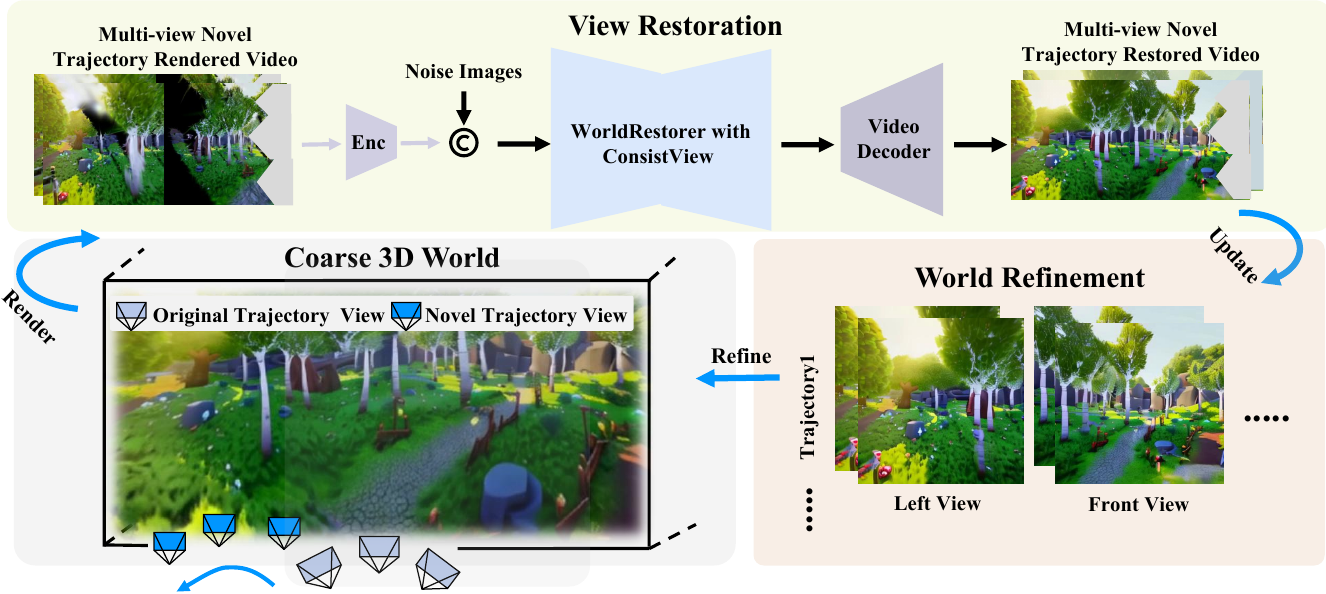}
    \caption{Overview of the WonderFree pipeline. WonderFree first builds a 3D world, then renders novel view videos along specified camera trajectories. WorldRestorer, equipped with the ConsistView mechanism, is applied to restore corrupted views while ensuring temporal coherence and spatial consistency. The restored videos are then used to refine the 3D world. This iterative process continues until the 3D world reaches a user-satisfactory level of visual quality and consistency.}
    \label{fig:WonderFree} 
\end{figure*}

\section{Related Work}

\subsection{ 3D World Generation} 
The generation of 3D scenes from a single image has been explored by various methods~\cite{text2room,luciddreamer,genex,pano2room,dreamscene360,diffpano,wonderworld,wonderturbo,diffusion360}. Some approaches first generate multiple views or panoramas of the scene and then convert them into 3D representations. For example, methods such as Text2Room~\cite{text2room} and LucidDreamer~\cite{luciddreamer} start with a single input image and a user-provided textual description to generate multi-view images and construct a 3D world. On the other hand, methods like GenEx~\cite{genex}, Pano2Room~\cite{pano2room}, and DreamScene360~\cite{dreamscene360} use pre-trained text-to-panoramic diffusion models~\cite{diffpano,diffusion360} to synthesize coherent panoramas, which are then upscaled to create a 3D world that can be explored. Meanwhile, to enhance the interactivity of generated scenes, WonderWorld~\cite{wonderworld} progressively constructs 3D worlds, allowing users to specify the content of the expanded 3D world through text. WonderTurbo~\cite{wonderturbo}, which builds upon WonderWorld~\cite{wonderworld}, achieves the generation of diverse scenes within 0.72 seconds by accelerating both geometric and appearance modeling, meeting the need for real-time interaction. 
However, these methods typically use a limited number of views to supervise the generation of 3D worlds, which constricts the explorability of the generated 3D environments. Specifically, these methods typically fail to render high-quality images when users attempt to move forward into unseen areas.

\subsection{Scene Reconstruction with Diffusion Prior} 
To enhance the explorability and spatial consistency of 3D worlds, many works~\cite{wonderworld,viewcrafter,wonderland,recondreamer,recondreamerplus,humandreamer,humandreamerx} have attempted to leverage generative models~\cite{svd,sd,cogvideo} to improve performance. WonderWorld~\cite{wonderworld} introduces a single-view inpainting approach aimed at uncovering and completing occluded regions in generated scene images using image inpainting techniques~\cite{sd}. However, inpainting from a single view cannot ensure consistency across multiple viewpoints, and image inpainting-based methods struggle to maintain spatial consistency. Wonderland~\cite{wonderland} adopts an approach based on the latent space of a video diffusion model~\cite{svd} to construct a 3D reconstruction framework, which utilizes a feedforward mechanism to predict 3D Gaussian Splatting~\cite{3dgs}. Although this method improves spatial consistency, the generated 3D worlds are limited in scale. Meanwhile, some works~\cite{viewcrafter,recondreamer,recondreamerplus,humandreamer,drivedreamer4d,sgd,3dgsenhancer} in other fields have also attempted to use generative models to further enhance the spatial consistency of 3D worlds, but these approaches are often restricted to one or two types of scenes, such as autonomous driving or indoor environments, whereas 3D world generation should encompass diverse scenarios. Additionally, these methods neglect consistency across multiple views. In contrast, WonderFree first constructs a video inpainting dataset containing various scenes through a dedicated pipeline, enabling WorldRestorer to effectively handle ghosting artifacts across diverse scenarios. Meanwhile, ConsistView is introduced to guarantee spatial consistency across multiple views.

\section{Method}
\subsection{Overview of the WonderFree framework.}
Current 3D world generation methods~\cite{wonderworld,text2room,luciddreamer,genex,pano2room,dreamscene360,diffpano,wonderturbo,diffusion360} often suffer from limited exploration capabilities due to supervision from only a limited number of views during scene generation. WonderWorld~\cite{wonderworld} adopts a single-view image inpainting approach, which neglects spatial consistency. Wonderland~\cite{wonderland} leverages video generation to improve temporal coherence within individual views but fails to maintain consistency between different viewpoints. In contrast, WonderFree markedly enhances visual quality and ensures both spatial and temporal consistency across viewpoints. 

As illustrated in Fig.~\ref{fig:WonderFree}, the pipeline of WonderFree begins by generating a coarse 3D world. To progressively refine this world, an iterative optimization loop is performed, comprising two steps: (1) view restoration, and (2) world  refinement. During view restoration, we render videos from new trajectories in the generated world. Due to the absence of supervision at these novel viewpoints during initial scene generation, the rendered videos often exhibit noticeable ghosting artifacts and distortions. Then, WorldRestorer, a diffusion-based restoration module, is applied to restore the corrupted novel views, conditioned on the rendered inputs. Crucially, to ensure spatial consistency, ConsistView is designed to jointly consider multiple viewpoints during the restoration process. In the subsequent world refinement stage, the restored videos serve as supervisory signals to iteratively improve both the geometric fidelity and multi-view consistency of the generated 3D world. This iterative loop continues until the resulting environment attains high visual quality and explorability.



\subsection{Training and Inference of WorldRestorer}
Traditional 3D world generation methods~\cite{wonderworld,text2room,luciddreamer,genex,pano2room,dreamscene360,diffpano,wonderturbo,diffusion360} often suffer from visual artifacts when users explore novel views or trajectories that lack sufficient supervision. Although recent approaches~\cite{wonderworld,wonderturbo,wonderjourney,sd} attempt to leverage powerful diffusion priors, most primarily rely on 2D priors and fail to effectively preserve spatial consistency across viewpoints. Additionally, video diffusion-based methods~\cite{wonderland,video1}, while capturing temporal coherence, cannot iteratively refine the underlying 3D representation. To address these limitations, we introduce WorldRestorer, a diffusion-based restoration module specifically designed to repair degraded renderings from novel views, while effectively maintaining spatial coherence and structural fidelity. In the following sections, we detail the training and inference procedures of WorldRestorer.

\noindent
\textbf{Training.}
A major challenge in training WorldRestorer lies in the absence of datasets specifically tailored for restoring artifacts arising from 3D world generation under novel explorations. To address this, we build a hybrid dataset of paired degraded and clean videos from synthetic and real environments (detled in ~\ref{sec:Dataset}). Based on the constructed dataset, WorldRestorer is implemented as a conditional video denoising network, which is initialized from a pretrained video generation model~\cite{svd}. During training, the degraded frames $\hat{V}$ are first masked with a spatial mask $M$, resulting in masked observations $\hat{V}_M = \hat{V} \odot M$, where $M$ is designed to highlight areas like occlusions. The masked inputs $\hat{V}_M$ are encoded via a video encoder $\mathcal{E}$:
\begin{equation}
\boldsymbol{z}_0 = \mathcal{E}(\hat{V}_M),
\end{equation}
and serve as the initial latent representation in a reverse diffusion process. We optimize the denoising network $\mathcal{D}$ using a conditional diffusion loss:
\begin{equation}
\mathcal{L}{_\mathcal{D}} = \mathbb{E}{_t, _\eta} 
\left[\left\|\eta-\mathcal{D}\left(\boldsymbol{z}_{n}, n, \boldsymbol{\mathcal{T}}\right)\right\|_{2}^{2}\right],
\end{equation}

where $\boldsymbol{z}_n$ is the latent representation corrupted by noise at step $n$, $\eta$ is the target noise to predict, and $\mathcal{T}$ denotes the degraded input frames as the control condition.

\begin{figure}[!t]
\centering
\setlength{\abovecaptionskip}{0.5em}
\includegraphics[width=0.99\textwidth]{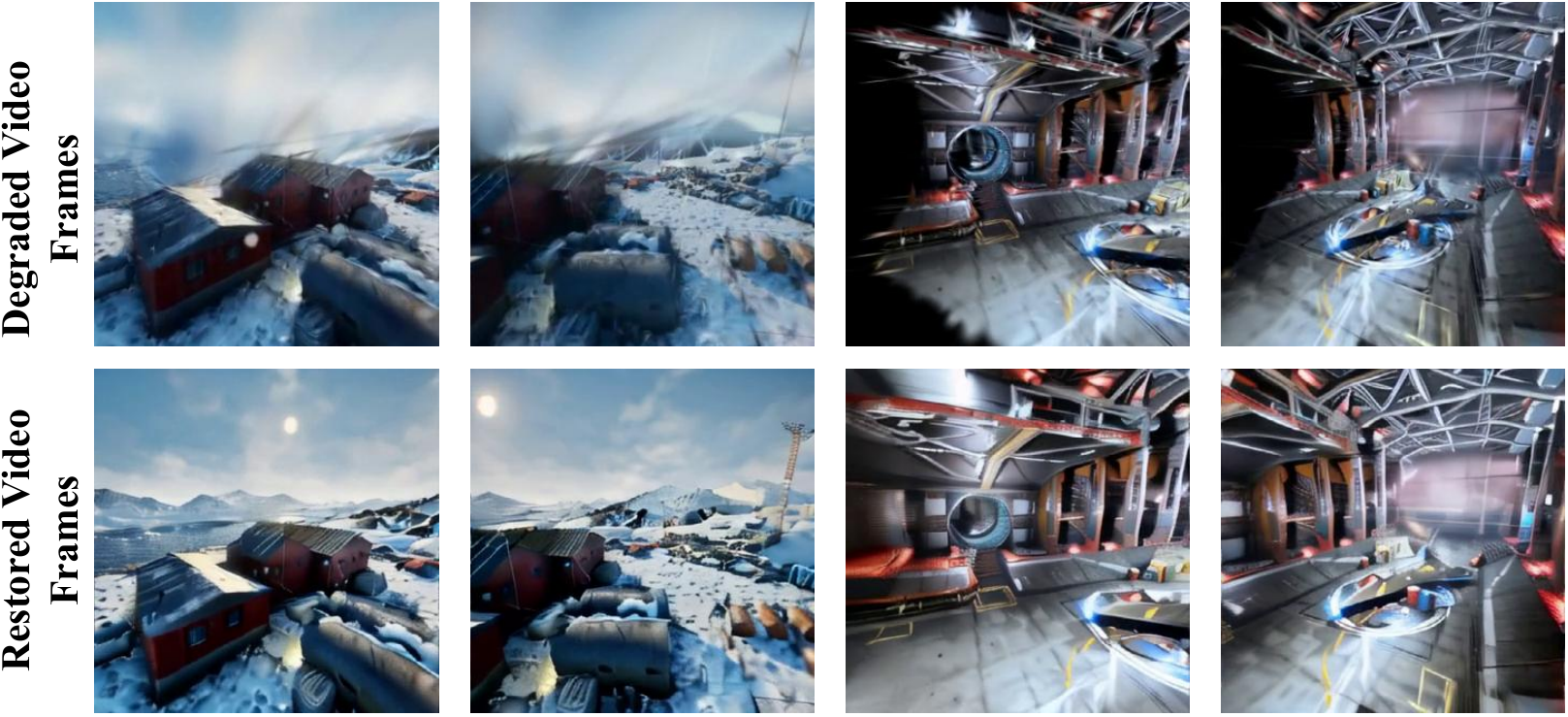}
\caption{Examples of degraded video frames and their restored counterparts. WorldRestorer with ConsistView not only eliminates floaters and artifacts but also maintains consistency across multiple views, including lighting effects..}
\label{fig:mitigate} 
\vspace{-1.5em}
\end{figure}

\noindent
\textbf{Inference.} After training WorldRestorer, we freeze its parameters and use it to refine the generated 3D world. We render videos along novel trajectories, then restore visual artifacts using the pretrained WorldRestorer. Subsequently, these restored videos are utilized to refine the 3D world, enhancing consistency and visual quality. As shown in Fig.~\ref{fig:mitigate}, the trained WorldRestorer effectively mitigates ghost artifacts.


\subsection{ConsistView}
\label{sec:ConsistView}
By constructing a video inpainting dataset and training a diffusion-based~\cite{sd,svd} video generation model~\cite{sd,svd,drivedreamer,drivedreamer2}, we can improve the exploration of 3D generated worlds. However, ensuring spatial consistency across multiple viewpoints remains a significant challenge. To mitigate this issue, existing methods~\cite{viewcrafter,mvsplat360,Evagaussians} leverage reference images as conditional inputs to guide 3D world construction. While this improves alignment with target views, it remains inadequate for ensuring consistency across all generated viewpoints.

Therefore, we propose ConsistView within the WorldRestorer framework. ConsistView is designed to unify spatial coherence within individual views and across views, which ensures temporal and spatial coherence. Specifically, in the constructed 3D generated world, we move along a continuous path at fixed intervals and capture $K$ different viewpoint images at each location simultaneously. These viewpoints are sampled with fixed angular offsets $\theta_k$ around the forward-facing direction, resulting in locally continuous and overlapping multi-view observations. Specifically, the angular offsets are defined as:
\begin{equation}
\theta_k = (k-n)\Delta\theta, \quad k = 0, 1, \ldots, 2n
\end{equation}
where $\Delta\theta$ represents the fixed angular step, and $n$ determines the total number of viewpoints on each side of the forward direction. Thus, $\theta_k$ spans the range from $-n\Delta\theta$ to $+n\Delta\theta$, corresponding to a total of $2n+1$ sampled viewpoints. Then, the camera rotation at the $k$-th viewpoint can be defined as:
\begin{equation}
R_k = R_0 \cdot \text{Rot}_y(\theta_k)
\end{equation}
where $R_0$ denotes the rotation matrix of the forward view, and $\text{Rot}_y(\theta_k)$ represents a rotation around the vertical axis. This process produces a group of images at each time step $t$, resulting in a multi-view video sequence $x = \{x_t^k\}_{t=1,k=1}^{T,K}$, where each image has spatial dimensions of $H \times W$. Then, we sequentially concatenate images from the multi-view video, aligning them with angular offsets $\theta_k$ in a clockwise direction. This results in a spatially unified image $x' \in \mathbb{R}^{T \times 3 \times H \times KW}$ at each time step. Then, the unified image $x'$ is fed into WorldRestorer to obtain a restored result that is consistent across all viewpoints. This strategy will be both employed during the training and inference stages. Contrary to processing each viewpoint separately, which may introduce visual inconsistencies, the unified image \( x' \) ensures consistent restoration across all viewpoints. This enhances coherence and user immersion. Additionally, dynamically adjusting \( K \) during training enables effective scalability and high-quality results for varying numbers of views.


\begin{figure}[!t]
\centering
\setlength{\abovecaptionskip}{0.5em}
\includegraphics[width=0.99\textwidth]{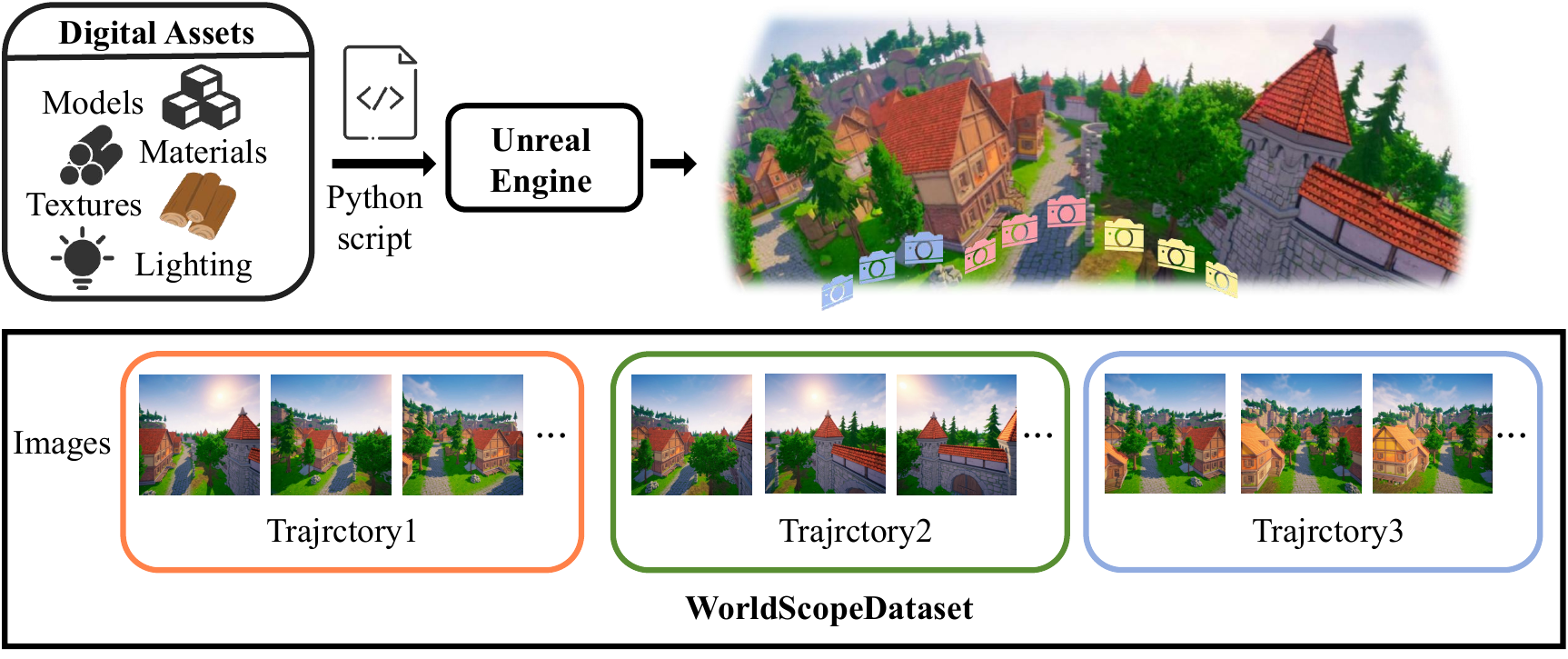}
\caption{The process of collecting data in Unreal Engine.}
\label{fig:dataset} 
\vspace{-1.5em}
\end{figure}

\begin{table}[t]
\centering
\caption{Detailed comparison with other mainstream 3D datasets. The scene types \crbx{crIndoor}{I}, \crbx{crNature}{N}, and \crbx{crUrban}{U} represent ``Indoor'', ``Nature'', and ``Urban'' scenes, respectively.}
\begin{tabular}{@{}lcccccc@{}}
\toprule
\textbf{Dataset} & \textbf{Year} & \textbf{Type} & \textbf{Source} & \textbf{\#Images} & \textbf{\#Scenes} & \textbf{Multi-view} \\ \midrule
KITTI~\cite{kitt}       & 2012          & \crbx{crUrban}{U}   & Real            & 15K               & 1                 & No                  \\
NYUv2~\cite{NYUv2}       & 2012          & \crbx{crIndoor}{I}  & Real            & 1.4K              & 464               & No                  \\
SceneNN~\cite{scenenn}     & 2016          & \crbx{crIndoor}{I}  & Real            & 502K              & -                 & No                  \\
ScanNet~\cite{scannet}     & 2017          & \crbx{crIndoor}{I}  & Real            & 2.5M              & -                 & No                  \\
Replica~\cite{replica}     & 2019          & \crbx{crIndoor}{I}  & Real            & 18                & -                 & No                  \\
nuScenes~\cite{nuscenes}    & 2020          & \crbx{crUrban}{U}   & Real            & 1.4M              & 2                 & No                  \\
Hypersim~\cite{hypersim}    & 2021          & \crbx{crNature}{I}  & Synthetic       & 77.4K             & 461               & No                  \\
3D-FRONT~\cite{3dfront}    & 2021          & \crbx{crNature}{I}  & Synthetic       & 6,813             & -                 & No                  \\
LHQ~\cite{LHQ}        & 2021          & \crbx{crNature}{N}  & Real            & 91.7K             & -                 & No                  \\
ACID~\cite{infinite}         & 2021          & \crbx{crNature}{N}  & Real            & 2.1M              & -                 & No                  \\
ScanNet++~\cite{scannetplus}   & 2023          & \crbx{crIndoor}{I}  & Real            & 11.1M             & 1006              & No                  \\
Our              & 2025          & \crbx{crIndoor}{I}\crbx{crUrban}{U}\crbx{crNature}{N}  & Real \& Synthetic & 23.4M             & 6523            & Yes                  \\ \bottomrule
\end{tabular}
\label{tab:our_dataset}
\end{table}


\subsection{WorldScopeDataset: Large-Scale and Multi-View  Dataset}
\label{sec:Dataset}
Crafting a model that directly restores corrupted novel views to enhance 3D world generation is intuitive, but collecting suitable training data makes it challenging. To address this, we propose the WorldScopeDataset, a hybrid data construction strategy that combines real and synthetic sources to build a diverse dataset containing multi-style, multi-view video sequences. Based on the WorldScopeDataset, we construct a dedicated video restoration dataset by rendering degraded videos from under-trained 3DGS, and pairing these renderings with their corresponding ground-truth frames. The WorldScopeDataset dataset will be publicly released. A detailed comparison with other mainstream 3D datasets~\cite{wen20253d,infinite,scannetplus,LHQ,3dfront,hypersim,scannet,scenenn,NYUv2,kitt} is presented in Tab.~\ref{tab:our_dataset}. 

For realistic scenes, we directly construct paired degraded and clean video sequences from real-world videos~\cite{videodataset1,videodataset2,videodataset3,videodataset4,videodataset5}. Specifically, we begin by extracting a sequence of camera extrinsics $\mathcal{C} = \{C_l\}_{l=1}^L$ and sparse point clouds using COLMAP~\cite{colmap}, where $L$ denotes the total number of frames and $C_l$ represents the camera extrinsic parameters at frame $l$. To mimic the sparse-view supervision typically encountered in 3D world generation, we select frames at fixed intervals for 3DGS training. To capture varying reconstruction quality, we save $T$ model checkpoints at random intervals during training, resulting in under-trained models $\{\phi^{(t)}\}_{t=1}^T$. Each saved model $\phi^{(t)}$ is used to render a video along the original trajectory $\mathcal{C}$, producing a degraded video sequence:
\begin{equation}
\tilde{V}^{(t)} = \mathcal{R}{\phi^{(t)}}(\mathcal{C}),
\end{equation}
where $\mathcal{R}{\phi^{(k)}}$ denotes the rendering function under model $\phi^{(k)}$. To simulate multiple viewpoints, each rendered video $\tilde{V}^{(k)}$ is divided into $N$ equal-length segments, with each segment representing a distinct virtual view along the trajectory. This strategy satisfies the training requirements of ConsistView to enable the model to learn spatial consistency across different perspectives. Due to early-stage underfitting, the rendered frames $\tilde{V}^{(k)}$ typically exhibit noticeable artifacts such as ghosting, geometry holes, or floaters. The corrupted videos are then paired with the original high-quality video $V{_\text{gt}}$ to form the dataset ${D}_{\text{real}}$:
\begin{equation}
\mathcal{D}_{\text{real}} = \left\{ \left( \tilde{V}^{(t)}, V_{\text{gt}} \right) \right\}_{t=1}^T.
\end{equation}

For non-photorealistic scenes, we generate synthetic data using Unreal Engine (UE)~\cite{unreal}, which provides a wide variety of customizable scene components. By composing these assets through automated scripts, we efficiently create a large number of diverse and stylized environments. In each scene, we define diverse continuous camera trajectories to simulate user exploration of the generated 3D world as shown in Fig.~\ref{fig:dataset}. At regular intervals along the trajectory, multiple viewpoints are simultaneously recorded with fixed angular offsets relative to the forward-facing direction, ensuring overlapping fields of view between adjacent perspectives (detailed in ~\ref{sec:ConsistView}). Then, following the same data processing pipeline as for \(\mathcal{S}_{\text{real}}\), we construct a new dataset denoted as \(\mathcal{S}_{\text{synth}}\).

Finally, we merge $\mathcal{S}_{\text{synth}}$ and $\mathcal{S}_{\text{real}}$ to obtain $\mathcal{S}_{\text{all}}$. Training video restoration models on this multi-view dataset improves spatial consistency, making it well-suited for robust and coherent 3D scene generation.

\section{Experiments}


\begin{figure}[t]
  \centering
  \includegraphics[width=0.99\linewidth]{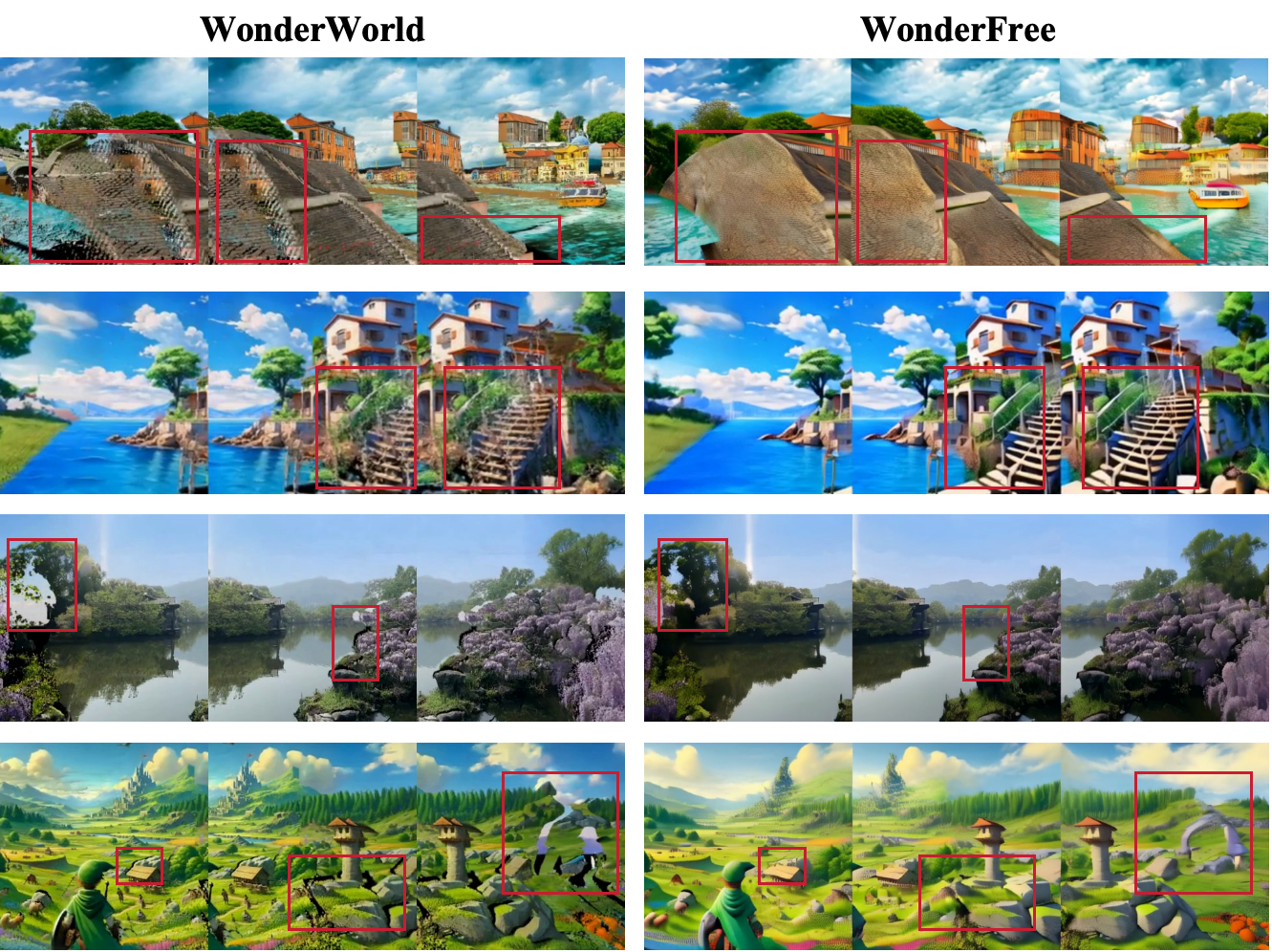} 
  \caption{Qualitative comparisons between WonderFree and WonderWorld~\cite{wonderworld} after large camera movements.}
  \label{fig:qualitative_comparison}
\end{figure}

\begin{table}[t]
\centering
\caption{Evaluation of novel view renderings for 3D world generation methods.} 
\label{tab:rgb_quality_comparison} 
\resizebox{\linewidth}{!}{
\begin{tabular}{lccccc} 
  \toprule
  Method & CLIP Score $\uparrow$ & CLIP Consistency $\uparrow$ & CIQA $\uparrow$ & Q-Align $\uparrow$ & CLIP Aesthetic $\uparrow$ \\ 
  \midrule
  LucidDreamer~\cite{luciddreamer} & 31.35 & 0.854 & 0.439 & 2.934 & 5.576 \\
  Text2Room~\cite{text2room} & 34.58 & 0.835 & 0.543 & 2.359 & 4.912 \\
  DreamScene360~\cite{dreamscene360} & 30.24 & 0.765 & 0.426 & 2.145 & 4.873 \\
  WonderJourney~\cite{wonderjourney} & 28.13 & 0.862 & 0.472 & 3.121 & 5.682 \\
  WonderWorld~\cite{wonderworld} & 32.28 & 0.913 & 0.560 & 3.437 & 6.123 \\
  WonderTurbo~\cite{wonderturbo} & 32.19 & 0.922 & 0.562 & 3.732 & 6.173 \\  
  \midrule 
  WonderFree  & \textbf{35.00} & \textbf{0.927} & \textbf{0.563} & \textbf{3.912} & \textbf{6.493} \\
  
  \bottomrule
\end{tabular}%
}
\end{table}

\begin{table}[t]   
    \centering
    \caption{Comparing the win rates of WonderFree in rendering novel views.}
    \footnotesize
    \label{tbl:user}
    \resizebox{\columnwidth}{!}{
    \setlength{\tabcolsep}{0.1cm}
    \begin{tabular}{cccccc}
    \toprule
    Method & Win Rate & Method & Win Rate & Method & Win Rate \\
    \midrule
    vs. LucidDreamer~\cite{luciddreamer} & 94.10\% & vs. Text2Room~\cite{text2room} & 93.70\% & vs. DreamScene360~\cite{pano2room} & 93.10\% \\
    vs. WonderJourney~\cite{dreamscene360} & 93.80\% & vs. WonderWorld~\cite{wonderjourney} & 77.20\% & vs. WonderTurbo~\cite{wonderworld} & 78.40\% \\
    \bottomrule
    \end{tabular}
    }
\end{table}

\subsection{Experimental Setup}

\textbf{Baselines.} 
For our comparative experiments, we select a diverse set of state-of-the-art 3D generation techniques~\cite{wonderworld,text2room,luciddreamer,pano2room,dreamscene360,wonderjourney}, including LucidDreamer~\cite{luciddreamer} and Text2Room~\cite{text2room} which synthesize 3D scenes through multi-view image generation, DreamScene360~\cite{dreamscene360} which produce panoramas subsequently reconstructed into 3D environments, and WonderJourney~\cite{wonderjourney}, WonderWorld~\cite{wonderworld} and WonderTurbo~\cite{wonderturbo} which generate 3D scenes through interactive methods.

\paragraph{Evaluation Metrics}
To evaluate the performance of 3D scene generation, we follow the evaluation protocol used in WonderWorld~\cite{wonderworld} and employ multiple quality assessment metrics, such as CLIP score~\cite{clip}, CLIP-based consistency, CLIP-IQA+\cite{clipiqa}, Q-Align\cite{qalin}, and CLIP aesthetic score. Additionally, we perform a user study to gather subjective feedback on visual quality. More comprehensive details can be found in the supplementary materials.

\paragraph{Implementation Details}
To ensure a thorough and unbiased evaluation, we source input images from three distinct methods: LucidDreamer~\cite{luciddreamer}, WonderJourney~\cite{wonderjourney}, and WonderWorld~\cite{wonderworld}. For each of the 4 specified test cases, 8 unique scenes are created, resulting in a total of 32 generated scenes for assessment. To guarantee consistency across comparisons, we utilize the same camera configuration throughout scene generation and evaluation. Meanwhile, we select WonderWorld~\cite{wonderworld} as our method to build a coarse 3D world. (More details can be seen in the supplementary materials.)

\subsection{Main Results}


\noindent
\textbf{Quantitative Results.}
In Tab.\ref{tab:rgb_quality_comparison}, we compare WonderFree with various state-of-the-art 3D scene generation methods\cite{wonderworld,text2room,luciddreamer,pano2room,dreamscene360,wonderjourney}. The experimental results indicate that interactive and progressively constructed methods exhibit significant advantages over techniques that rely on multi-view image synthesis~\cite{text2room,luciddreamer}. Notably, WonderWorld achieves superior performance across all metrics compared to these conventional approaches~\cite{text2room,luciddreamer,dreamscene360}. Moreover, WonderTurbo enhances performance further by leveraging a specialized dataset tailored explicitly for 3D scene generation. Furthermore, WonderFree improves scene construction by integrating video diffusion priors and ensuring multi-view consistency, making it the top-performing method across all metrics.

\noindent
\textbf{User Study.}
Additionally, we conduct a user study to evaluate the quality of 3D scenes generated by various methods. As shown in Tab.~\ref{tbl:user}, the results indicate that WonderFree achieves comparable performance to WonderWorld~\cite{wonderworld} and WonderTurbo~\cite{wonderturbo}.

\noindent
\textbf{Qualitative Results.} As shown in Fig. \ref{fig:qualitative_comparison}, we conduct a qualitative comparison between WonderFree and WonderWorld under the same novel view. Notably, after significant camera movements, WonderWorld exhibits substantial distortions and blurriness in several regions, such as inside the red box. However, WonderFree significantly mitigates these issues, enhancing image quality and improving explorability. Additionally, WonderFree maintains consistency across multiple viewpoints and reduces missing content caused by occluded regions.

\begin{figure}[!t]
\centering
\setlength{\abovecaptionskip}{0.5em}
\includegraphics[width=0.9\textwidth]{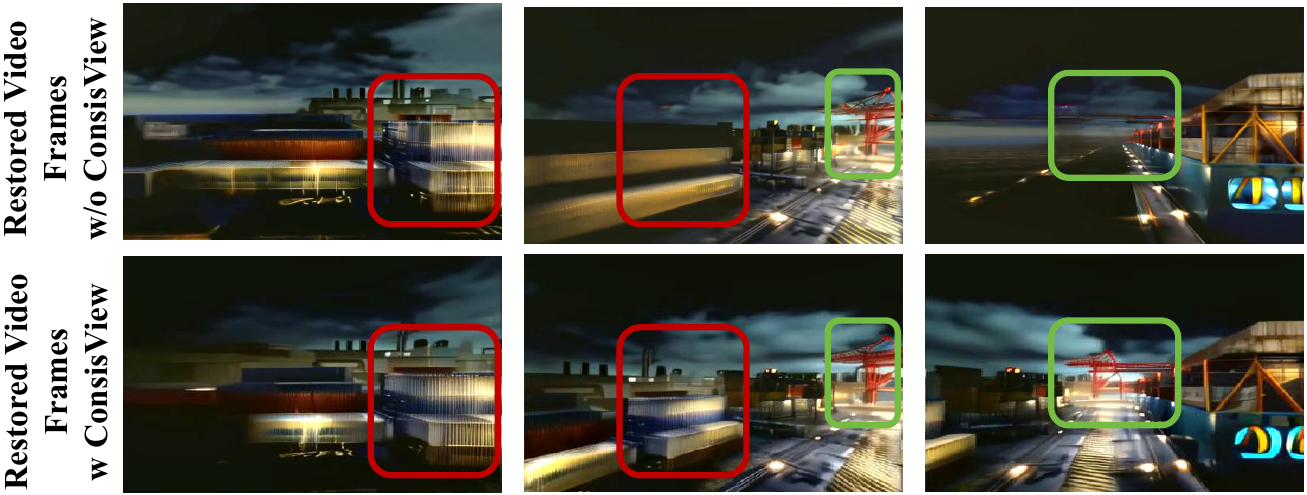}
\caption{Visualization comparison of WorldRestorer with and without ConsistView. The first, second, and third columns correspond to the left, front, and right views, respectively. Overlapping regions between views are indicated by the red and green boxes.}
\label{fig:multi} 
\vspace{-1.5em}
\end{figure}

\subsection{Ablation Studies}


\paragraph{WorldRestorer.}
As illustrated in Fig.~\ref{fig:mitigate}, we present examples comparing degraded video frames with their restored counterparts. These examples clearly demonstrate that our WorldRestorer combined with ConsistView effectively removes visual artifacts and floaters while preserving visual coherence, including consistent lighting and shading effects. As shown in Tab.~\ref{tab:ablation}, WorldRestorer alone notably improves the baseline performance across various evaluation metrics.

\paragraph{ConsistView.}
As shown in Fig.~\ref{fig:multi}, we compare WorldRestorer with and without ConsistView in a challenging low-light condition. The WorldRestorer without ConsistView, despite being capable of recovering missing content in each individual view and effectively removing floaters, struggles to ensure consistency across multiple views. Meanwhile, we compare the metrics of WorldRestorer with and without ConsistView, as shown in Tab.~\ref{tab:ablation}, the WorldRestorer with ConsistView achieves better performance in all metrics.


  

\begin{table}[t]
\centering
\caption{Evaluation of novel view renderings for 3D world generation methods.}
\label{tab:ablation}
\resizebox{\linewidth}{!}{%
\begin{tabular}{ccccccc}
\toprule
WorldRestorer & ConsistView & CLIP Score $\uparrow$ & CLIP Consistency $\uparrow$ & CIQA $\uparrow$ & Q-Align $\uparrow$ & CLIP Aesthetic $\uparrow$ \\
\midrule
 &  & 32.28 & 0.913 & 0.560 & 3.437 & 6.123 \\
\checkmark &  & 33.34 & 0.921 & 0.562 & 3.541 & 6.133 \\
\checkmark & \checkmark & \textbf{35.00} & \textbf{0.927} & \textbf{0.563} & \textbf{3.912} & \textbf{6.493} \\
\bottomrule
\end{tabular}%
}
\end{table}

\section{Conclusion}
Despite recent progress in 3D scene generation from a single image, explorability remains limited due to inefficiencies when navigating beyond original viewpoints. To address this, we decouple the challenge into two key sub-problems: novel view quality and cross-view consistency. Therefore, we propose WonderFree, an efficient framework that integrates video diffusion priors and emphasizes cross-view consistency to mitigate these limitations. Specifically, for improving novel view quality, we introduce WorldRestorer, a restoration model capable of effectively eliminating visual artifacts and distortions. To enhance cross-view consistency, we propose ConsistView, a multi-view joint restoration mechanism promoting spatial coherence across viewpoints. Moreover, for training WorldRestorer, we build the dataset WorldScopeDataset, which contains large and diverse scenes with multi-view data. Experimental results demonstrate that WonderFree significantly improves both rendering quality and overall consistency, facilitating immersive exploration of 3D scenes.

\normalem
\bibliographystyle{unsrt}
\bibliography{ref}

\newpage
\appendix  
\title{
WonderFree: Enhancing Novel View Quality and Cross-View Consistency for 3D Scene Exploration
}

\maketitle

\section{The Visualizations of the Video Restoration Dataset}
\begin{figure}[h]
\centering
\setlength{\abovecaptionskip}{0.5em}
\includegraphics[width=0.99\textwidth]{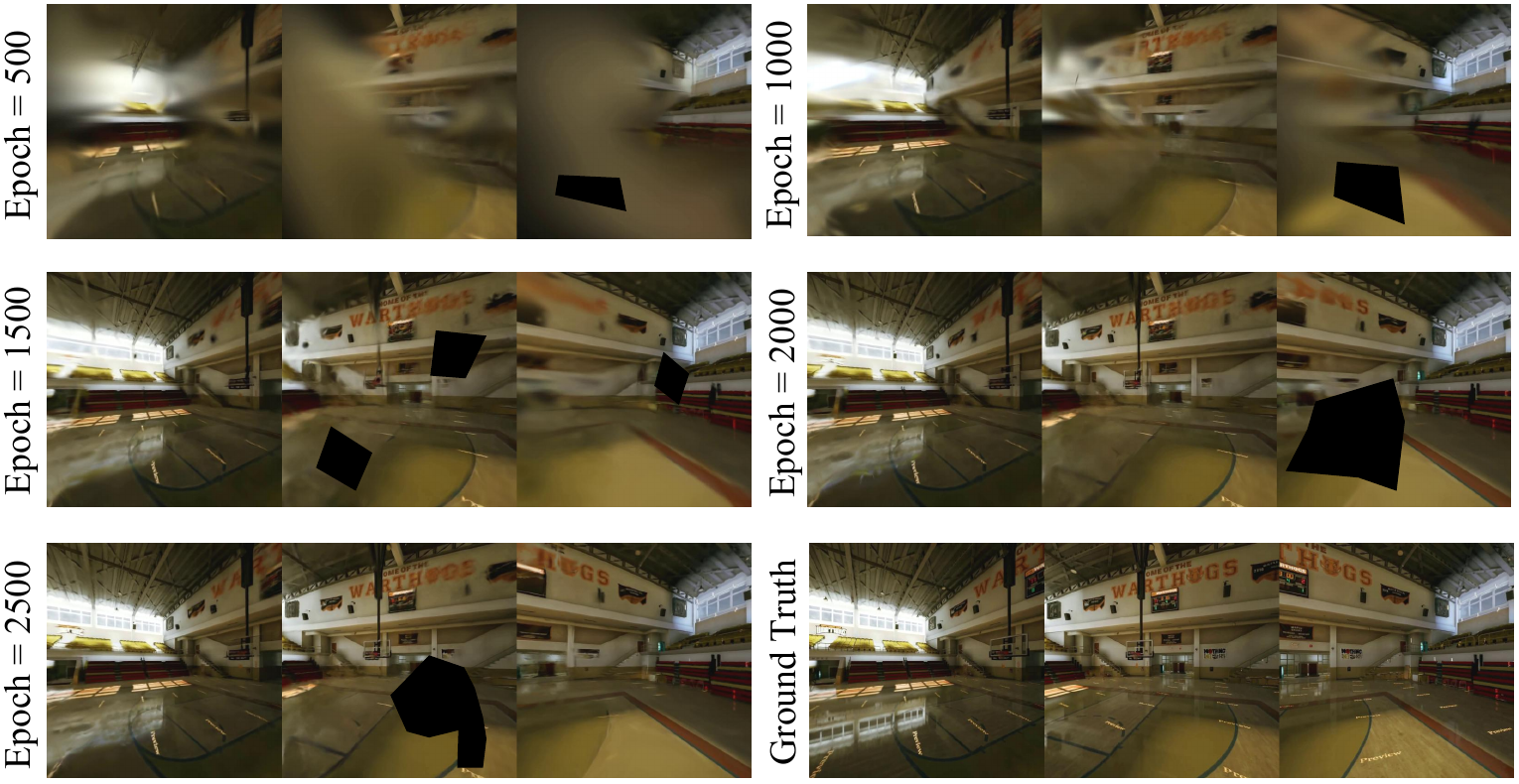}
\includegraphics[width=0.99\textwidth]{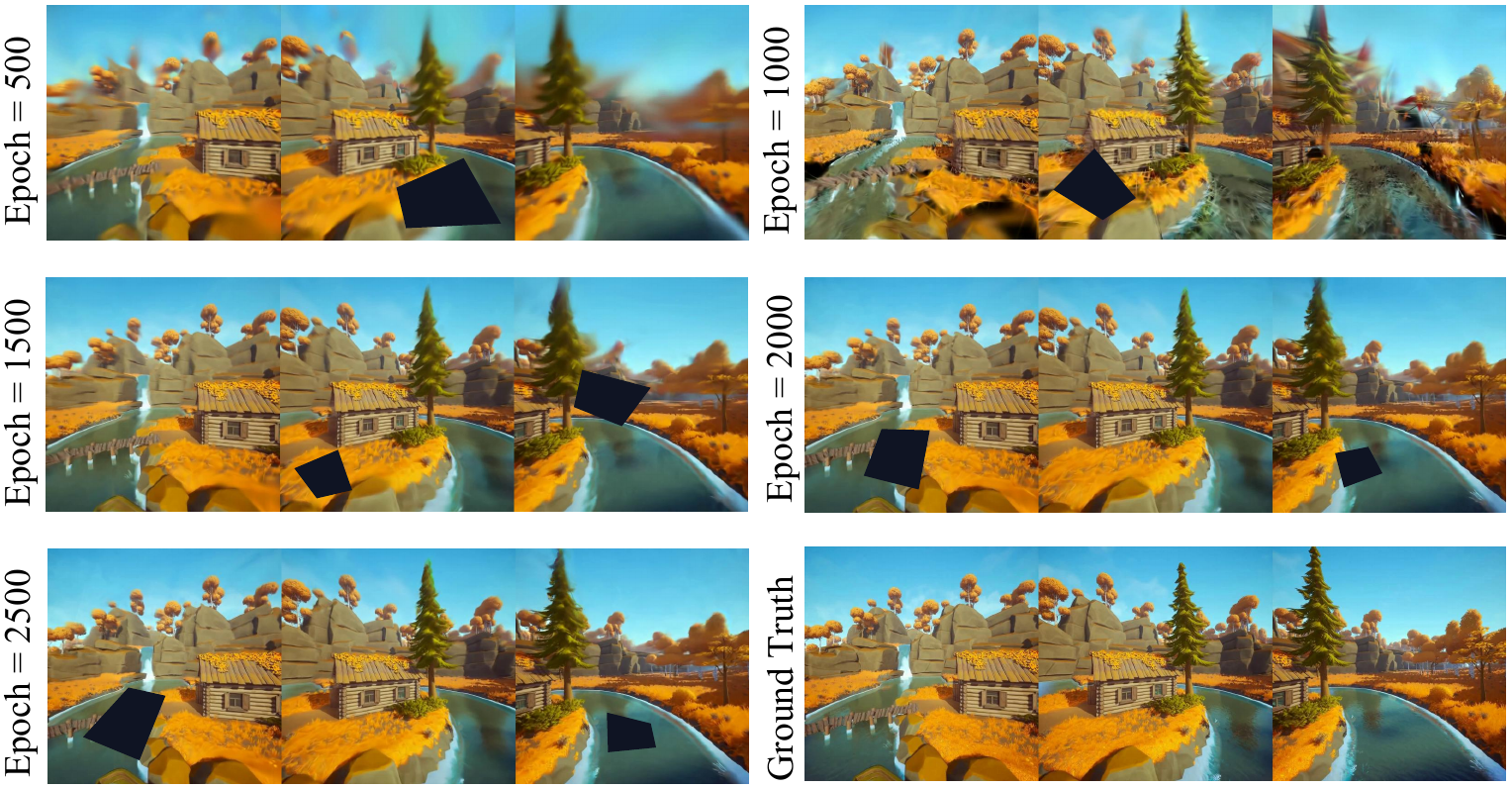}
\caption{The video restoration dataset for WorldRestorer training, where images rendered by 3DGS at various training epochs are individually paired with corresponding Ground Truth images. Artificial masks are introduced to simulate occlusions.}
\label{fig:re1} 
\vspace{-1.5em}
\end{figure}

As shown in Fig.~\ref{fig:re1}, we provide visualizations of the video restoration dataset tailored for 3D world generation, based on the WorldScopeDataset. Specifically, we collect rendered video clips from under-trained 3DGS reconstruction models, which inherently exhibit significant ghosting artifacts due to model underfitting. These degraded renderings are paired with their corresponding ground truth images to form training pairs. Additionally, to satisfy the multi-view training requirements of ConsistView, we simultaneously render scenes from multiple viewpoints, enabling the WorldRestorer to effectively learn spatial consistency across different perspectives. Moreover, to realistically simulate image degradation caused by occlusions, we artificially introduce masks onto the degraded rendered images. As illustrated in Fig.~\ref{fig:re1}, different types of rendering defects appear at distinct training stages. Specifically, significant blur artifacts occur during the early training stages (around 500–1000 epochs), while noticeable ghosting artifacts and detail loss are prominent at mid-training stages (1500–2500 epochs). By approximately 2500 epochs, although the overall quality of rendered images improves, subtle artifacts and blurriness remain, especially in fine details such as text. Since these defects frequently co-occur during novel view restoration, it is essential to gather examples containing all types simultaneously.

\section{Implementation Details}
\label{submission}

\noindent \textbf{Metrics.} To assess the quality of 3D world generation, several metrics such as the CLIP score~\cite{clip}, CLIP consistency~\cite{clip}, CLIP-IQA+~\cite{clipiqa}, Q-Align~\cite{qalin}, and the CLIP aesthetic~\cite{clip} score are utilized to evaluate both semantic alignment and visual aesthetics. Specifically, the CLIP score~\cite{clip} measures semantic relevance by computing the cosine similarity between CLIP embeddings of the input textual scene descriptions and the rendered images. CLIP consistency~\cite{clip} evaluates semantic coherence across various viewpoints by comparing their respective embeddings with the embedding of a central reference view. CLIP-IQA+~\cite{clipiqa} combines perceptual image quality assessment methods with deep learning techniques to deliver a comprehensive evaluation of overall image quality. Q-Align~\cite{qalin} assesses visual quality by training large multimodal models to predict human-aligned, discrete quality ratings, enhancing interpretability and generalization. Lastly, the CLIP aesthetic~\cite{clip} score assesses the visual attractiveness of rendered scenes by considering critical aesthetic elements such as composition, contrast, and color harmony.

\noindent \textbf{Camera Trajectories.} In most 3D scene generation methods~\cite{wonderworld,text2room,luciddreamer,genex,pano2room,dreamscene360,diffpano,wonderturbo,diffusion360}, simple panoramic camera paths are commonly used to evaluate the quality of generated worlds. However, these paths often do not sufficiently capture interactive movements typically performed by users during actual exploration. Therefore, we design more complex camera trajectories to better simulate realistic exploration scenarios. In addition to the conventional panoramic camera paths, we introduce five additional camera movements: camera moves forward, forward-left, forward-right, translates left, and translates right. During testing, we select views at regular intervals along each trajectory as novel views, and the final evaluation metrics are computed by averaging results across all paths. 
\section{Qualitative Results}
\begin{figure}[t]
\centering
\setlength{\abovecaptionskip}{0.5em}
\includegraphics[width=0.99\textwidth]{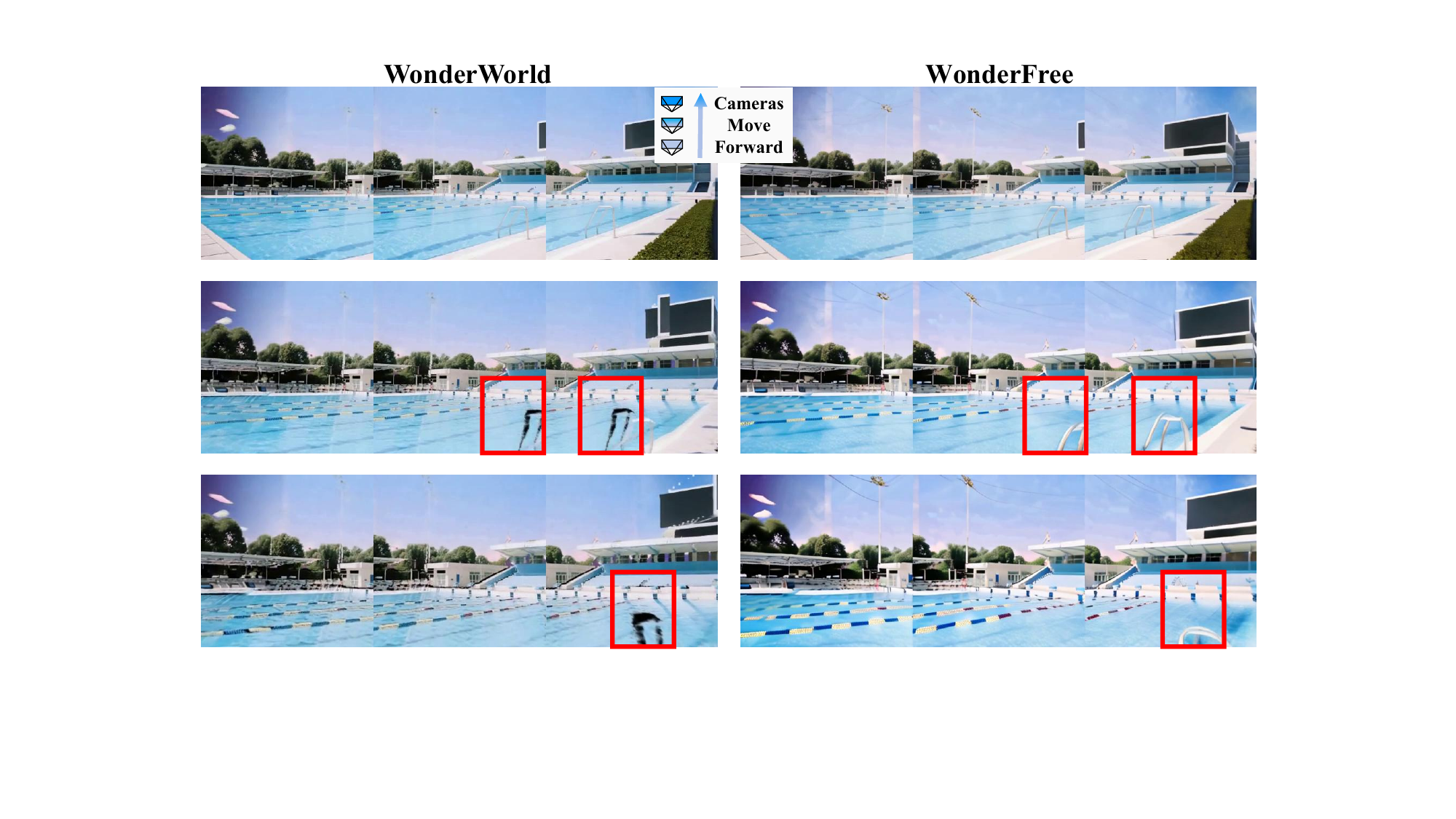}
\includegraphics[width=0.99\textwidth]{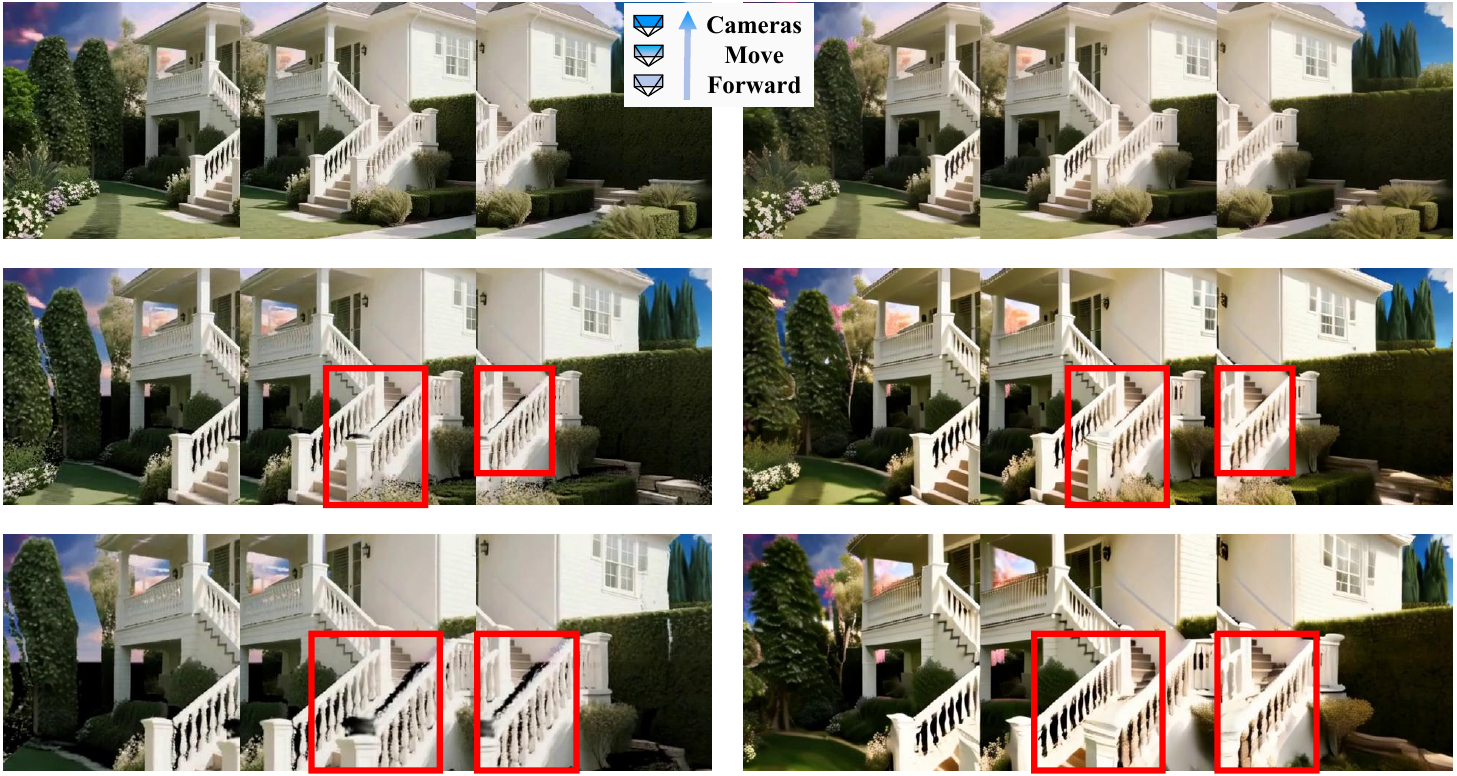}
\includegraphics[width=0.99\textwidth]{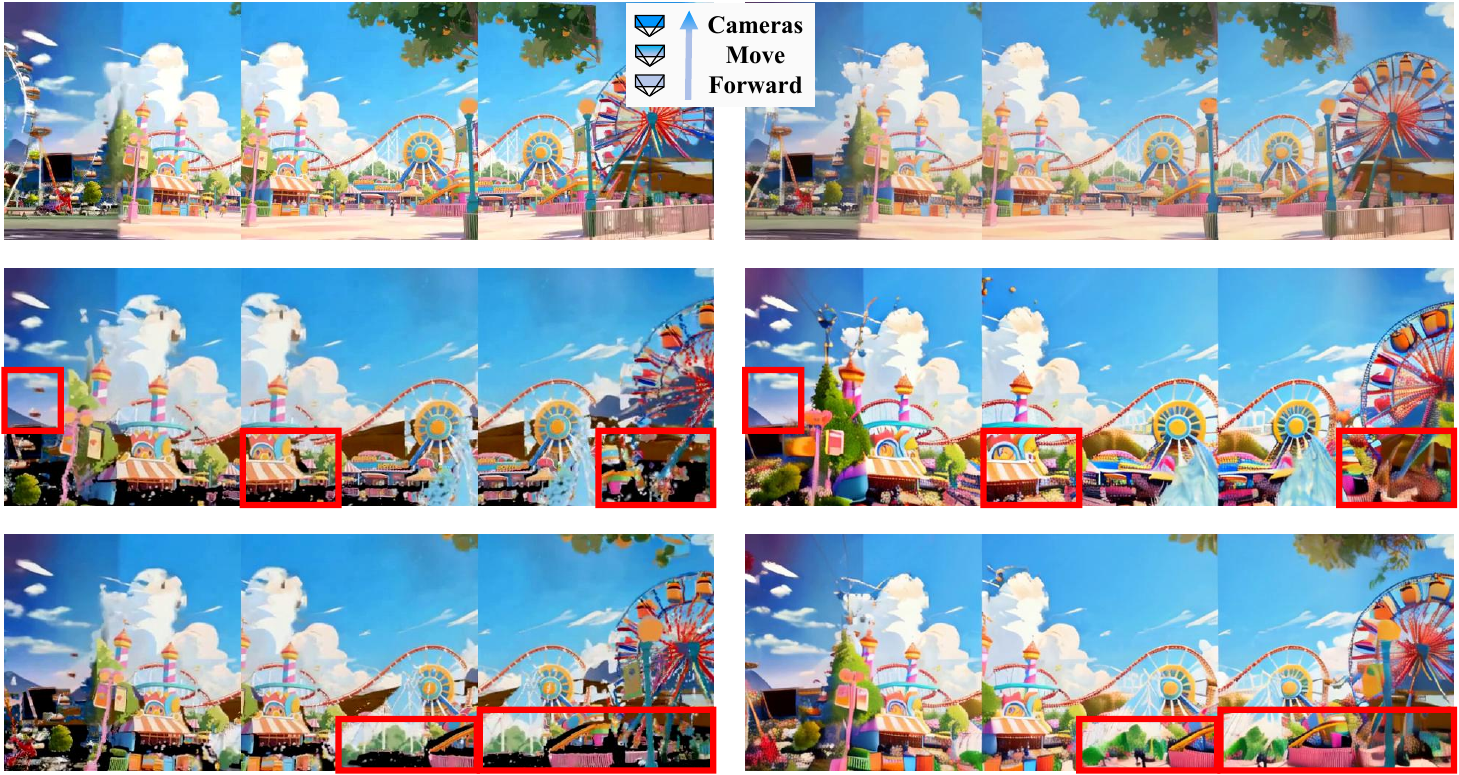}
\caption{Qualitative examples.}
\label{fig:q1} 
\vspace{-1.5em}
\end{figure}

\begin{figure}[t]
\centering
\setlength{\abovecaptionskip}{0.5em}
\includegraphics[width=0.99\textwidth]{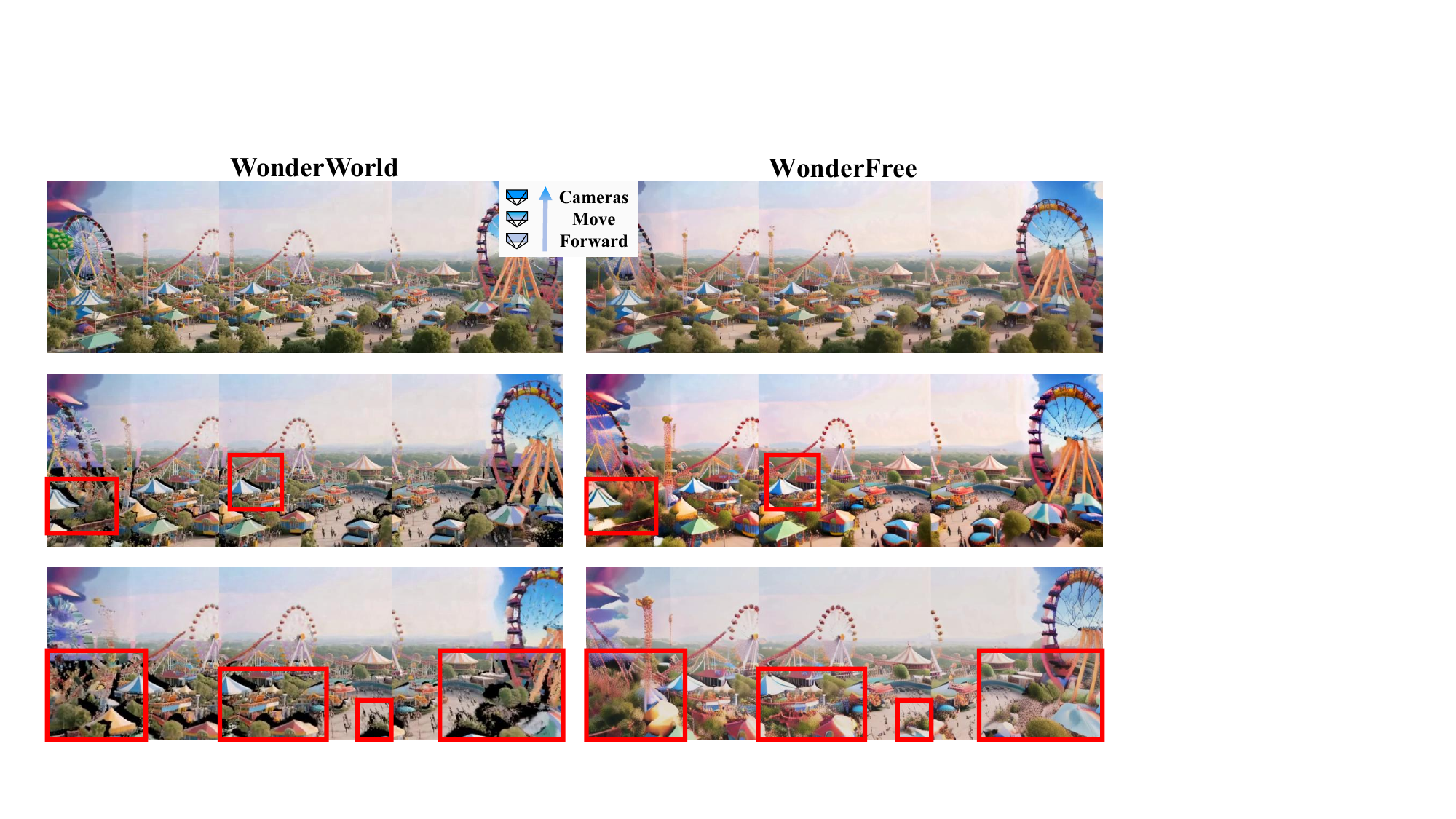}
\includegraphics[width=0.99\textwidth]{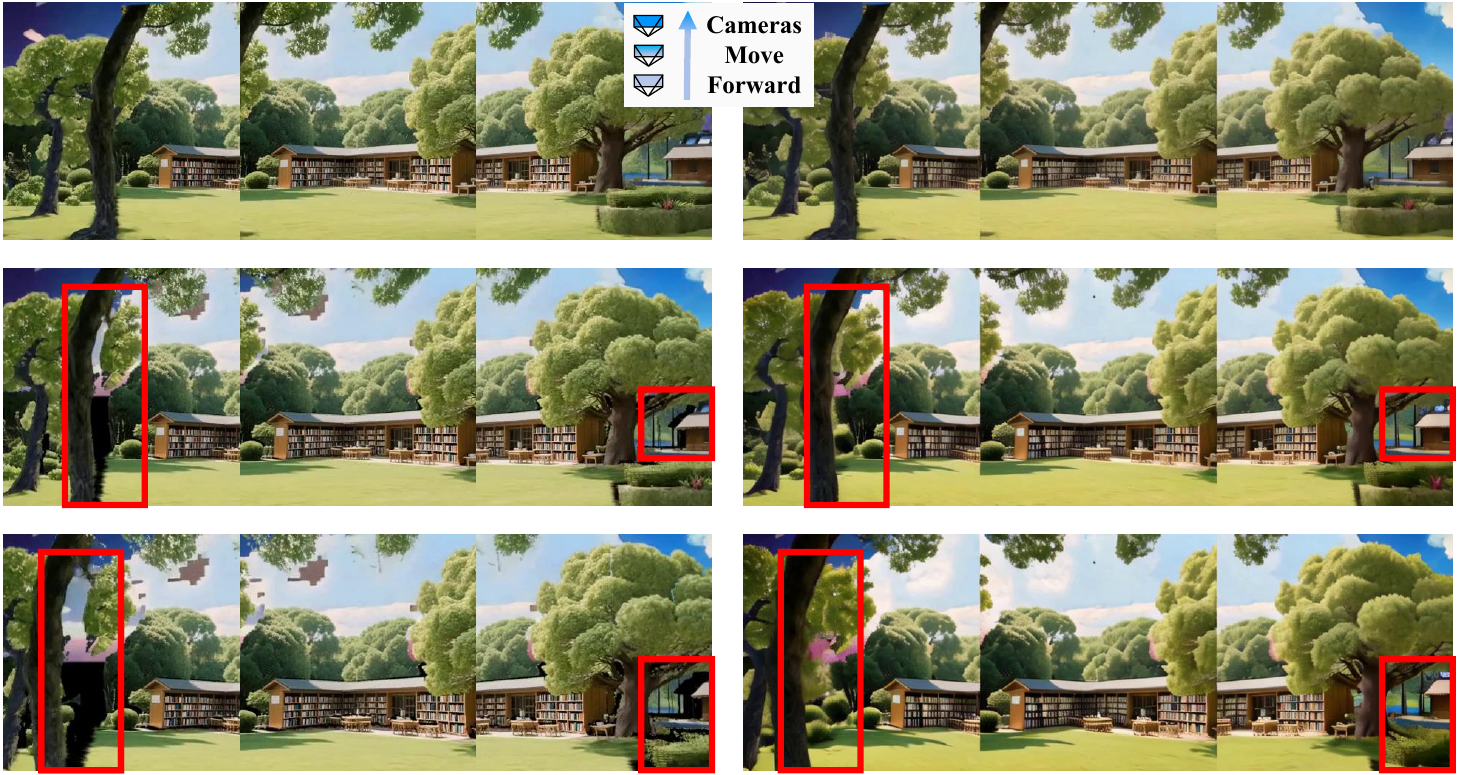}
\includegraphics[width=0.99\textwidth]{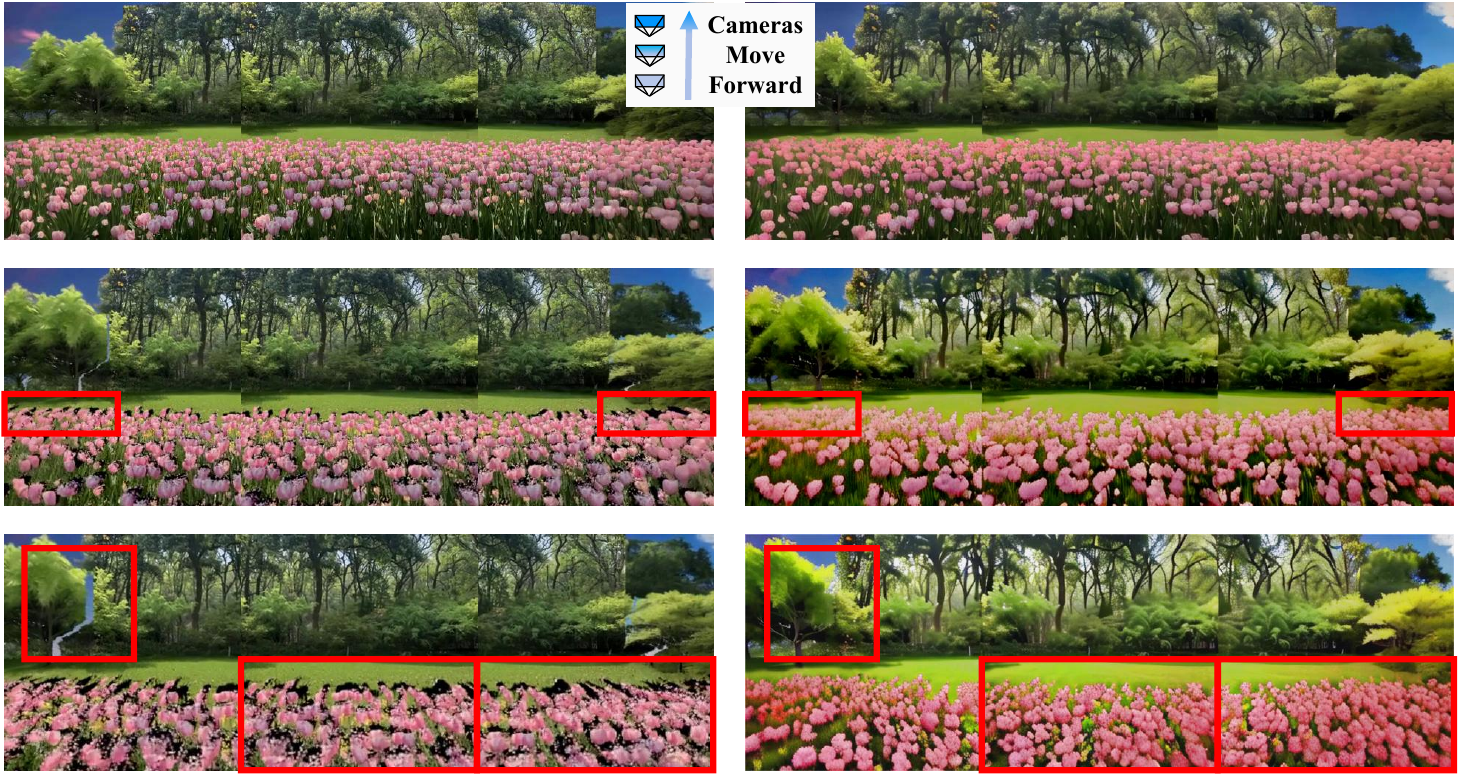}
\caption{Qualitative examples.}
\label{fig:q2} 
\vspace{-1.5em}
\end{figure}

\begin{figure}[t]
\centering
\setlength{\abovecaptionskip}{0.5em}
\includegraphics[width=0.99\textwidth]{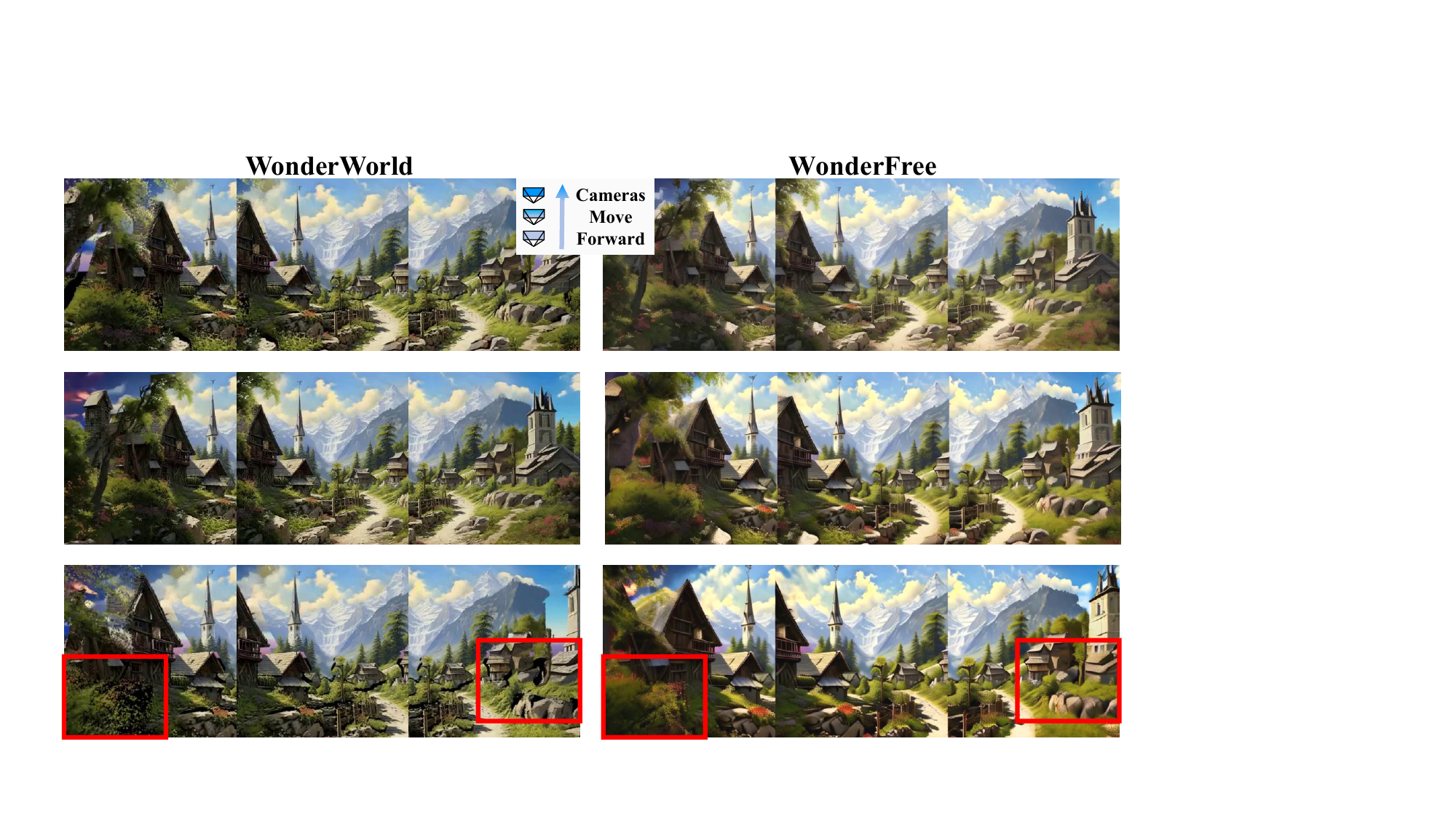}
\includegraphics[width=0.99\textwidth]{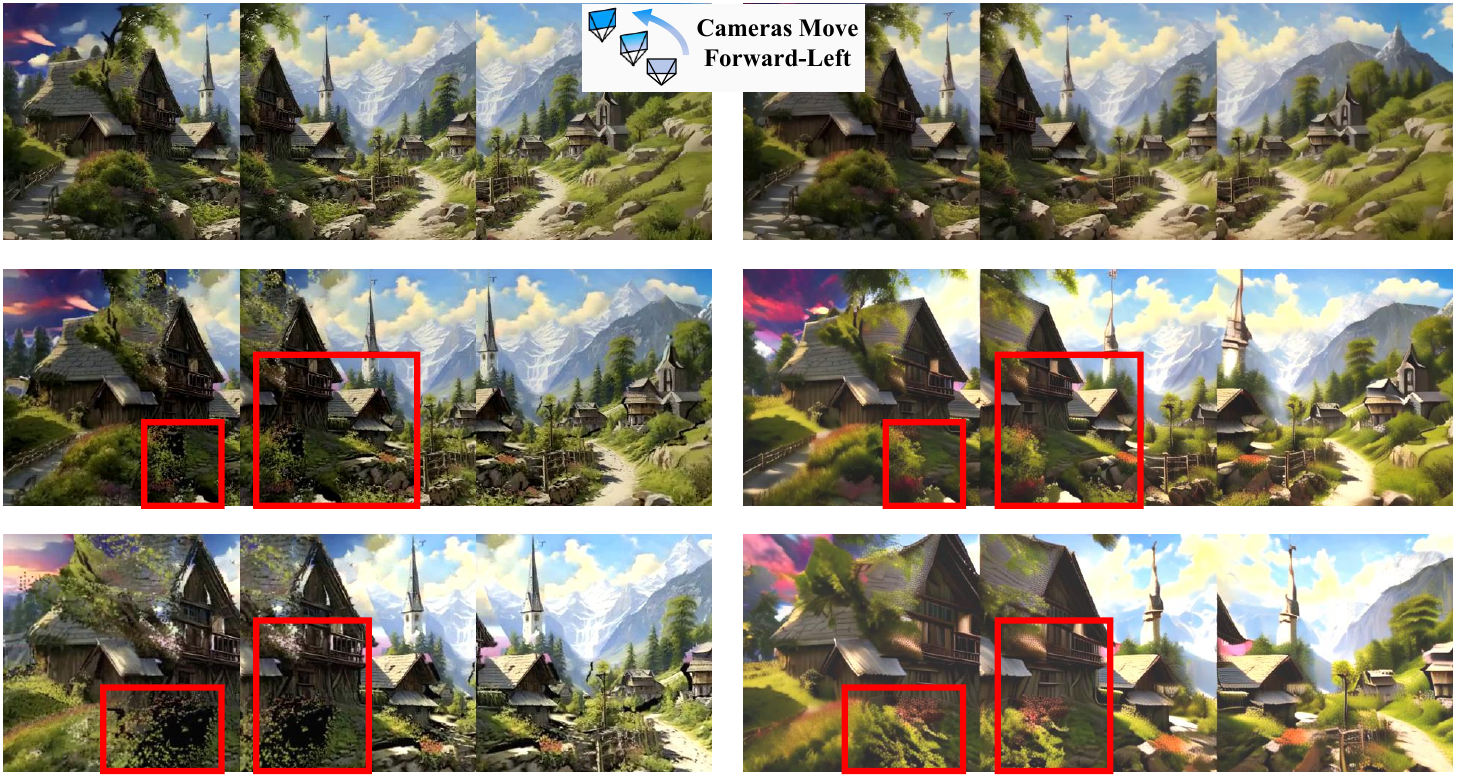}
\includegraphics[width=0.99\textwidth]{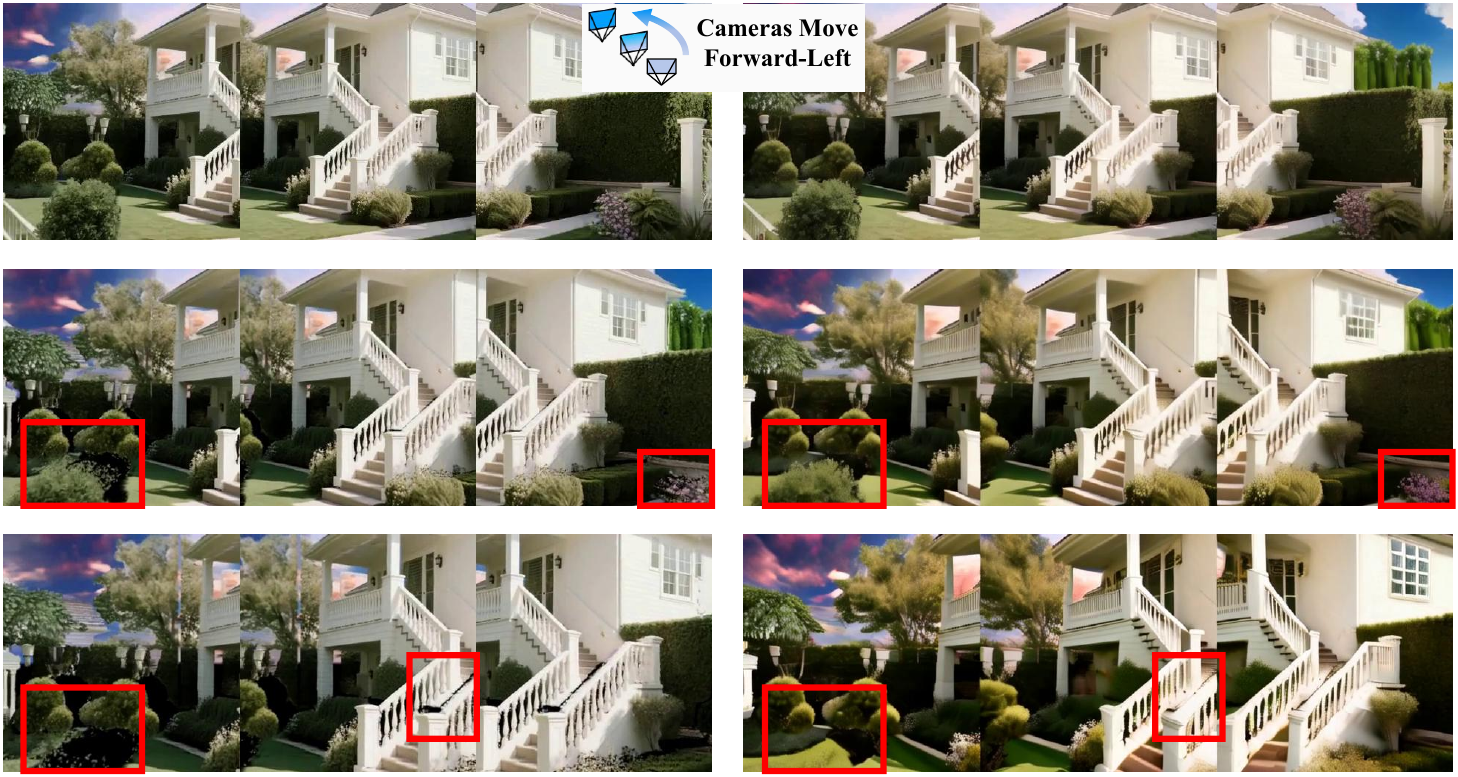}
\caption{Qualitative examples.}
\label{fig:q3} 
\vspace{-1.5em}
\end{figure}

\begin{figure}[t]
\centering
\setlength{\abovecaptionskip}{0.5em}
\includegraphics[width=0.99\textwidth]{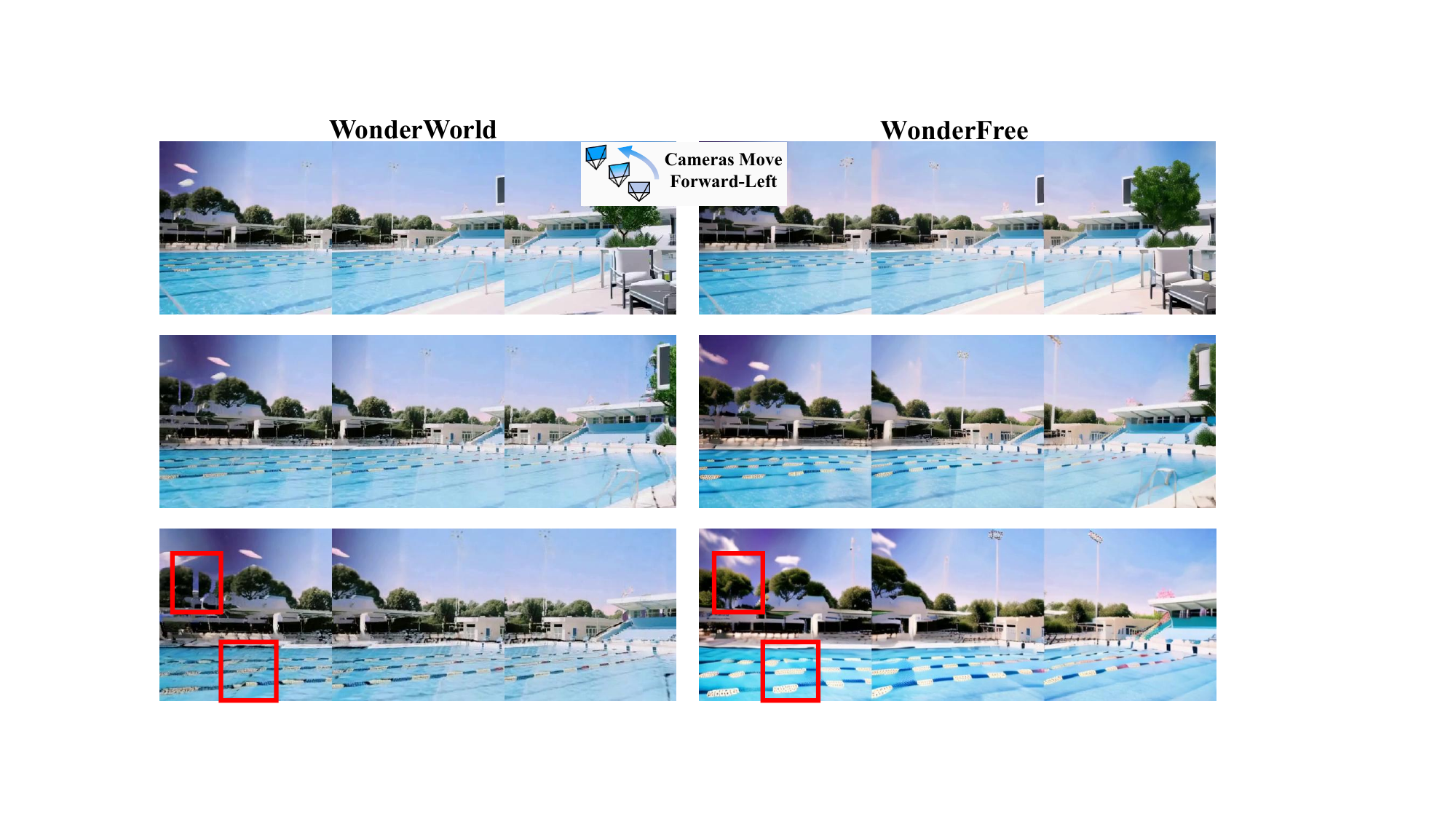}
\includegraphics[width=0.99\textwidth]{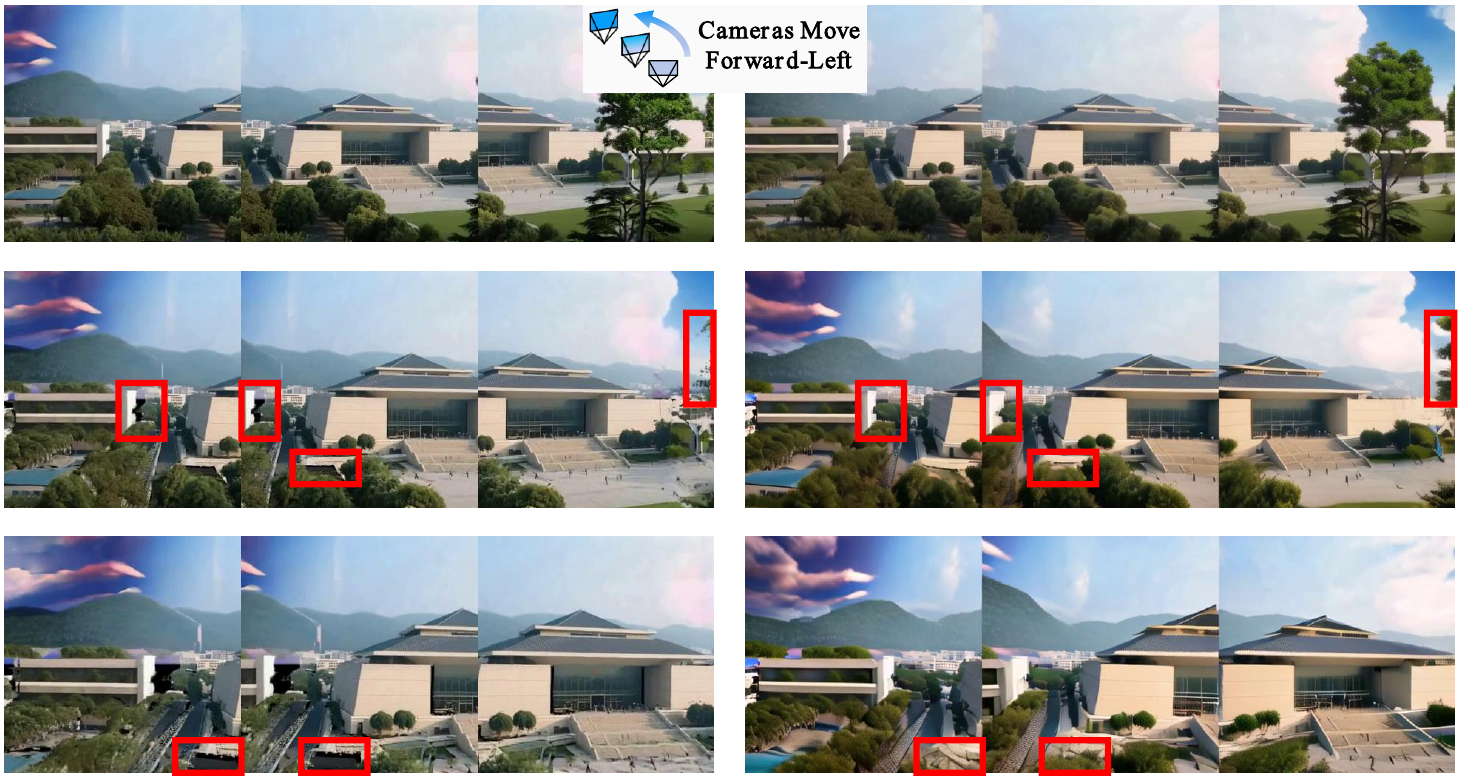}
\includegraphics[width=0.99\textwidth]{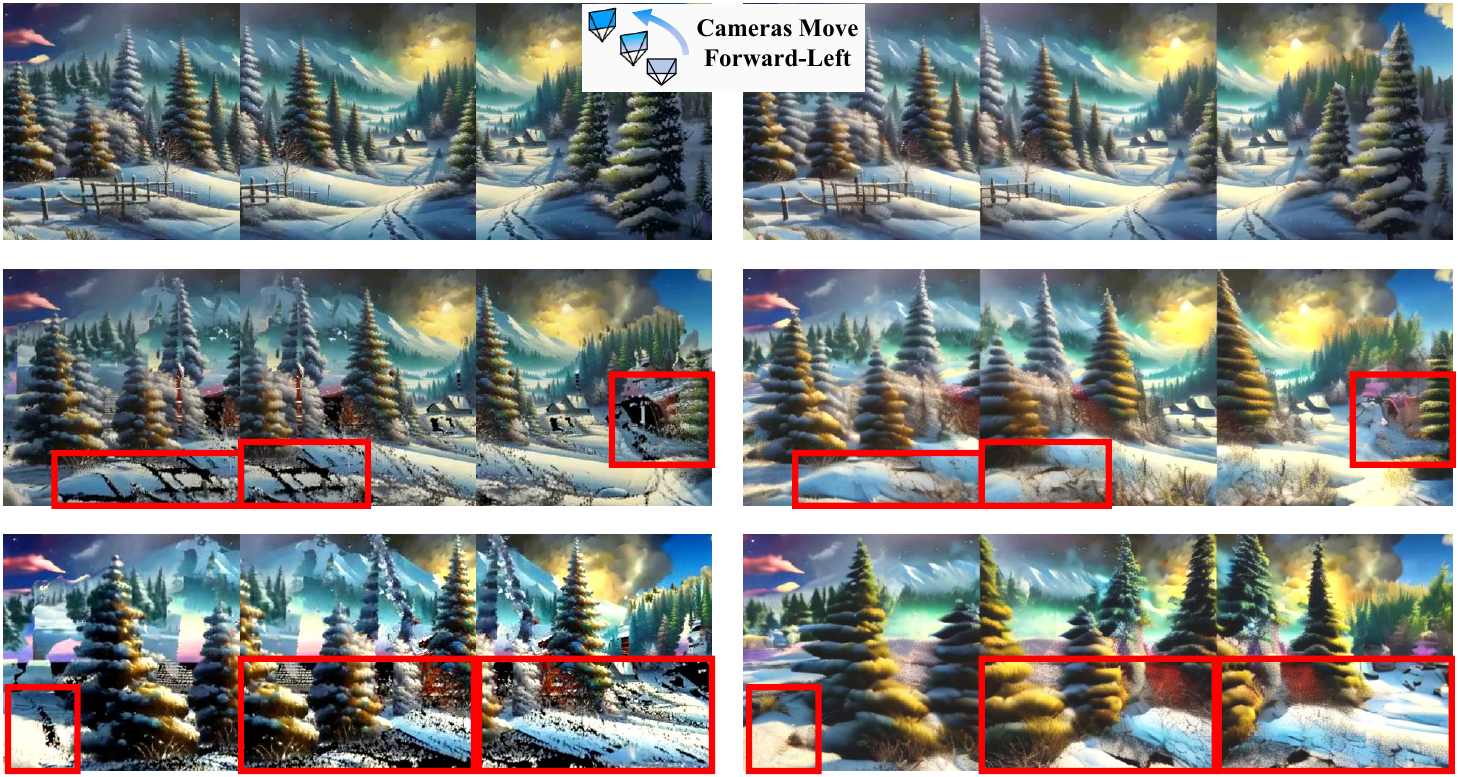}
\caption{Qualitative examples.}
\label{fig:q4} 
\vspace{-1.5em}
\end{figure}

\begin{figure}[t]
\centering
\setlength{\abovecaptionskip}{0.5em}
\includegraphics[width=0.99\textwidth]{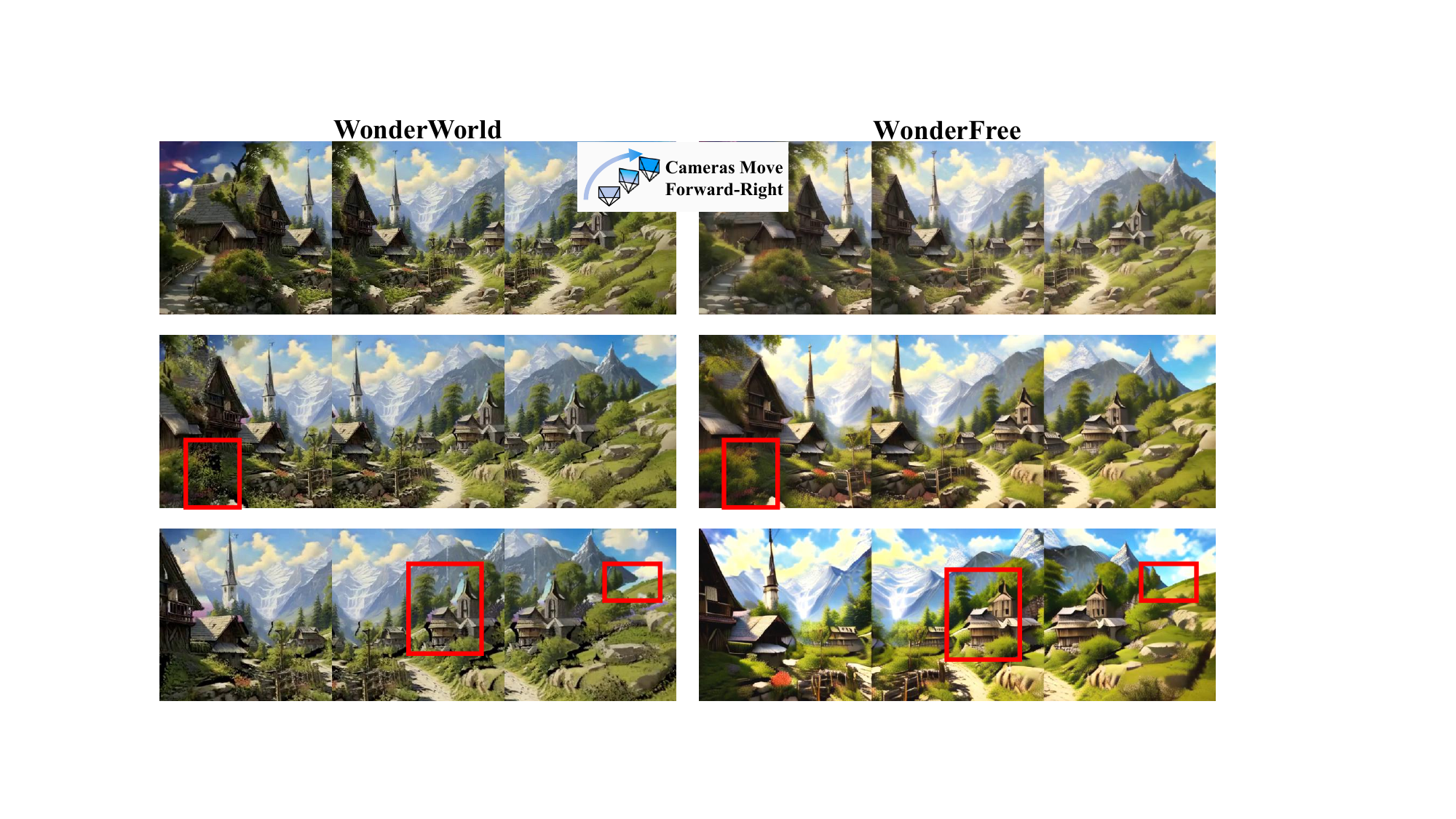}
\includegraphics[width=0.99\textwidth]{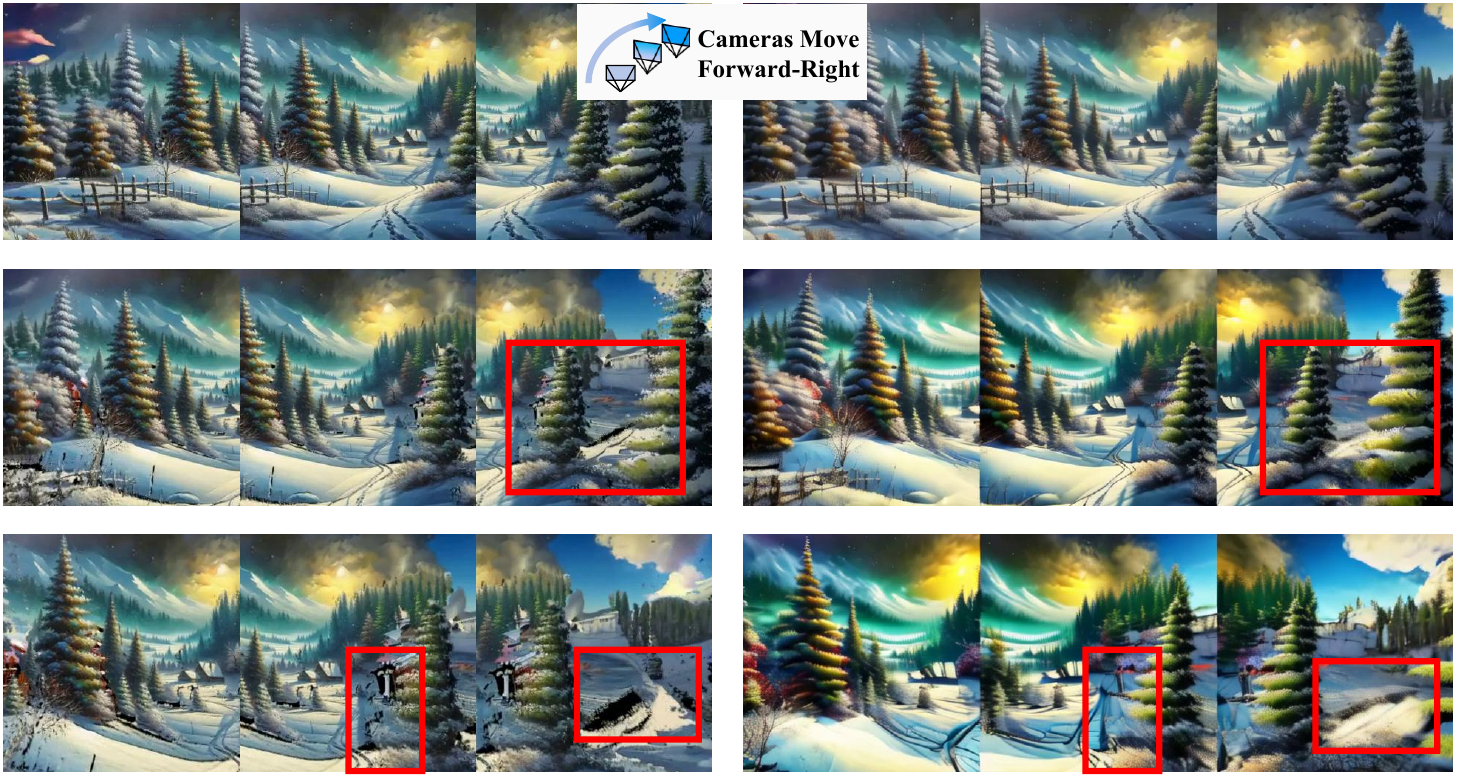}
\includegraphics[width=0.99\textwidth]{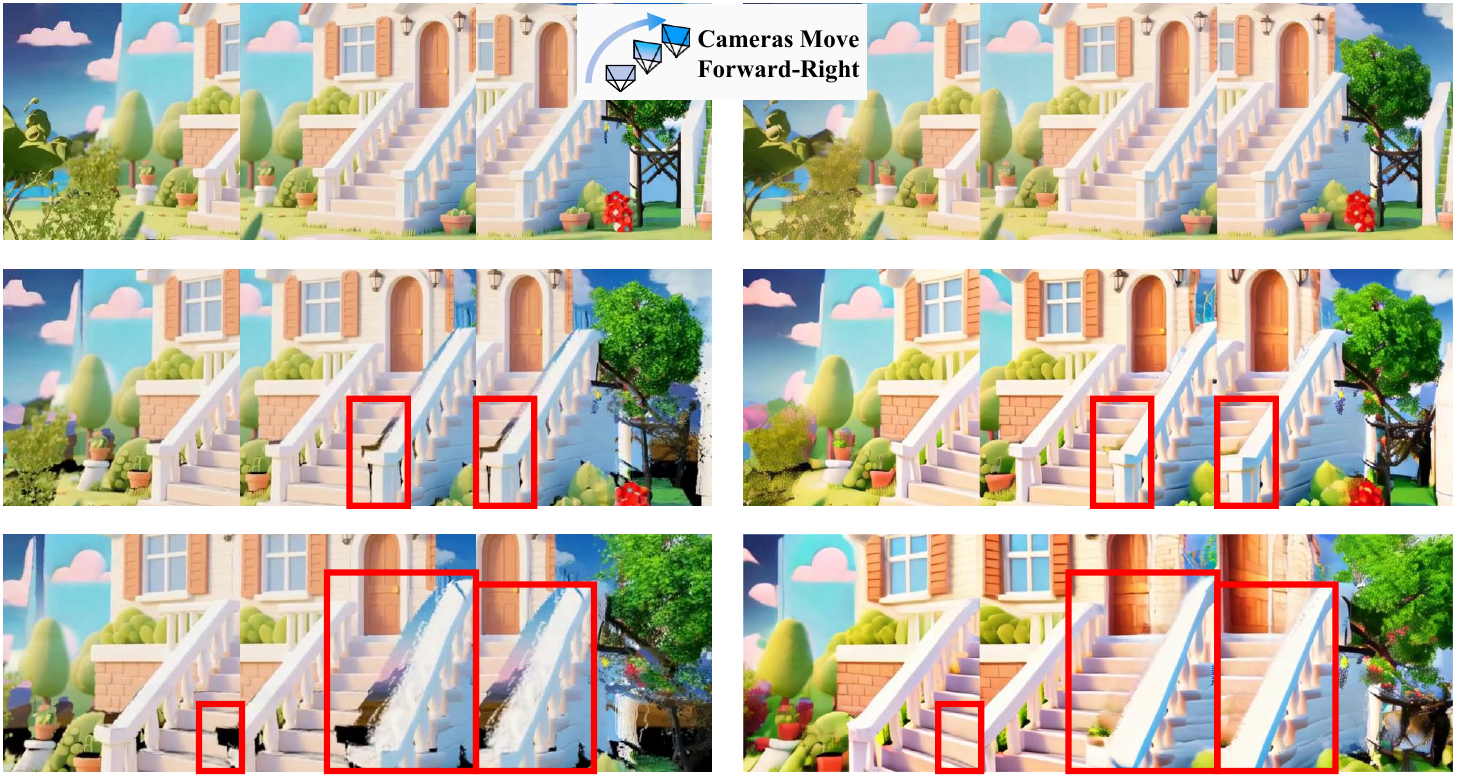}
\caption{Qualitative examples.}
\label{fig:q5} 
\vspace{-1.5em}
\end{figure}

\begin{figure}[t]
\centering
\setlength{\abovecaptionskip}{0.5em}
\includegraphics[width=0.99\textwidth]{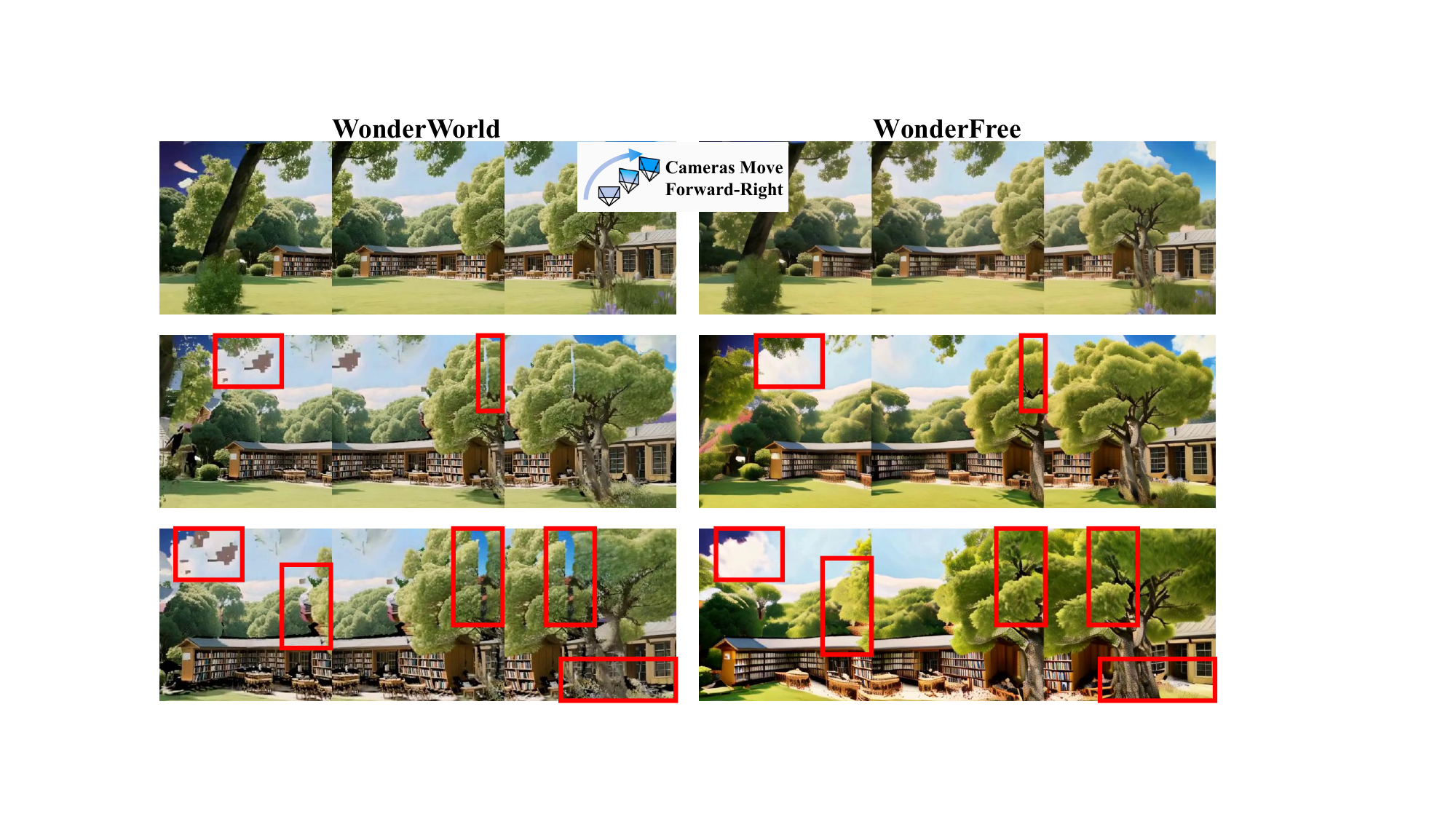}
\includegraphics[width=0.99\textwidth]{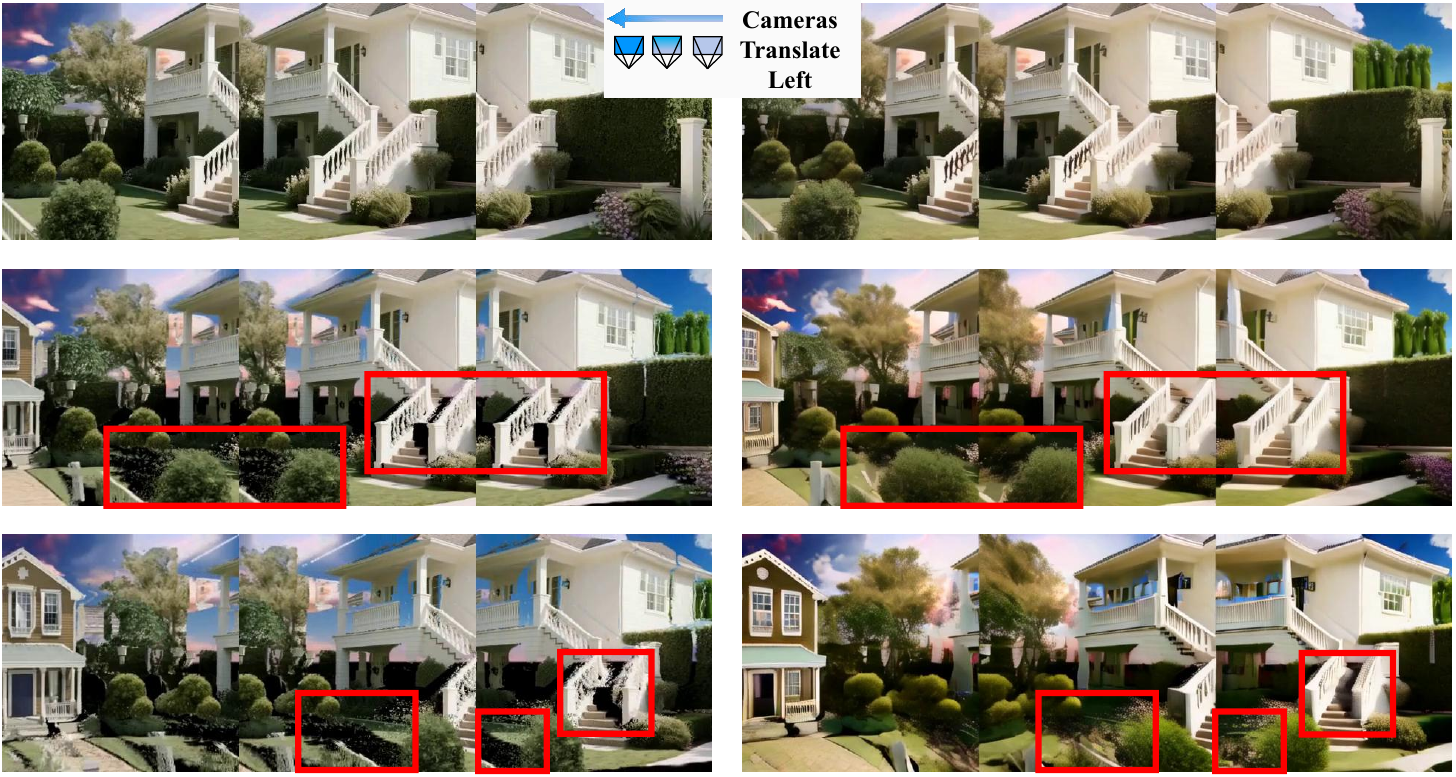}
\includegraphics[width=0.99\textwidth]{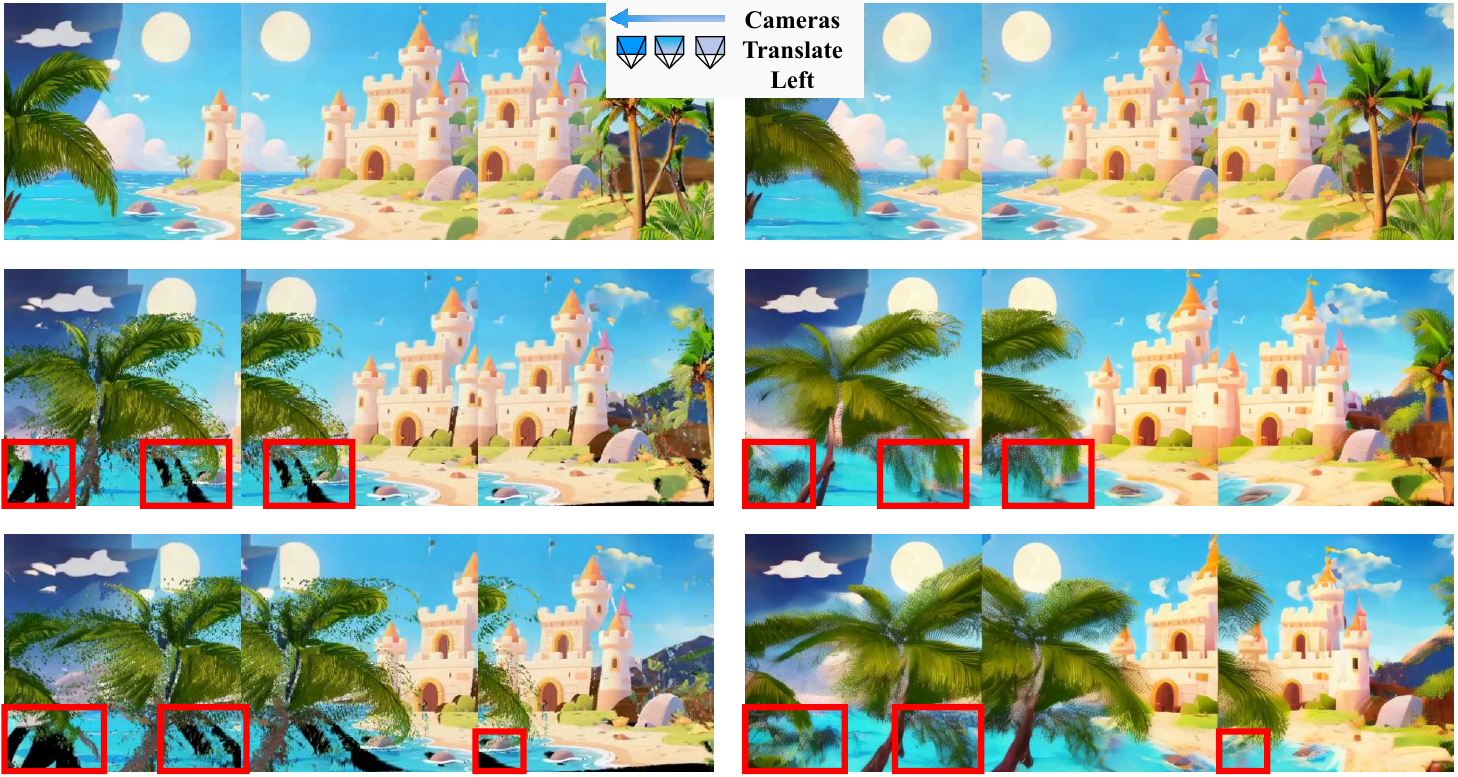}
\caption{Qualitative examples.}
\label{fig:q6} 
\vspace{-1.5em}
\end{figure}

\begin{figure}[t]
\centering
\setlength{\abovecaptionskip}{0.5em}
\includegraphics[width=0.99\textwidth]{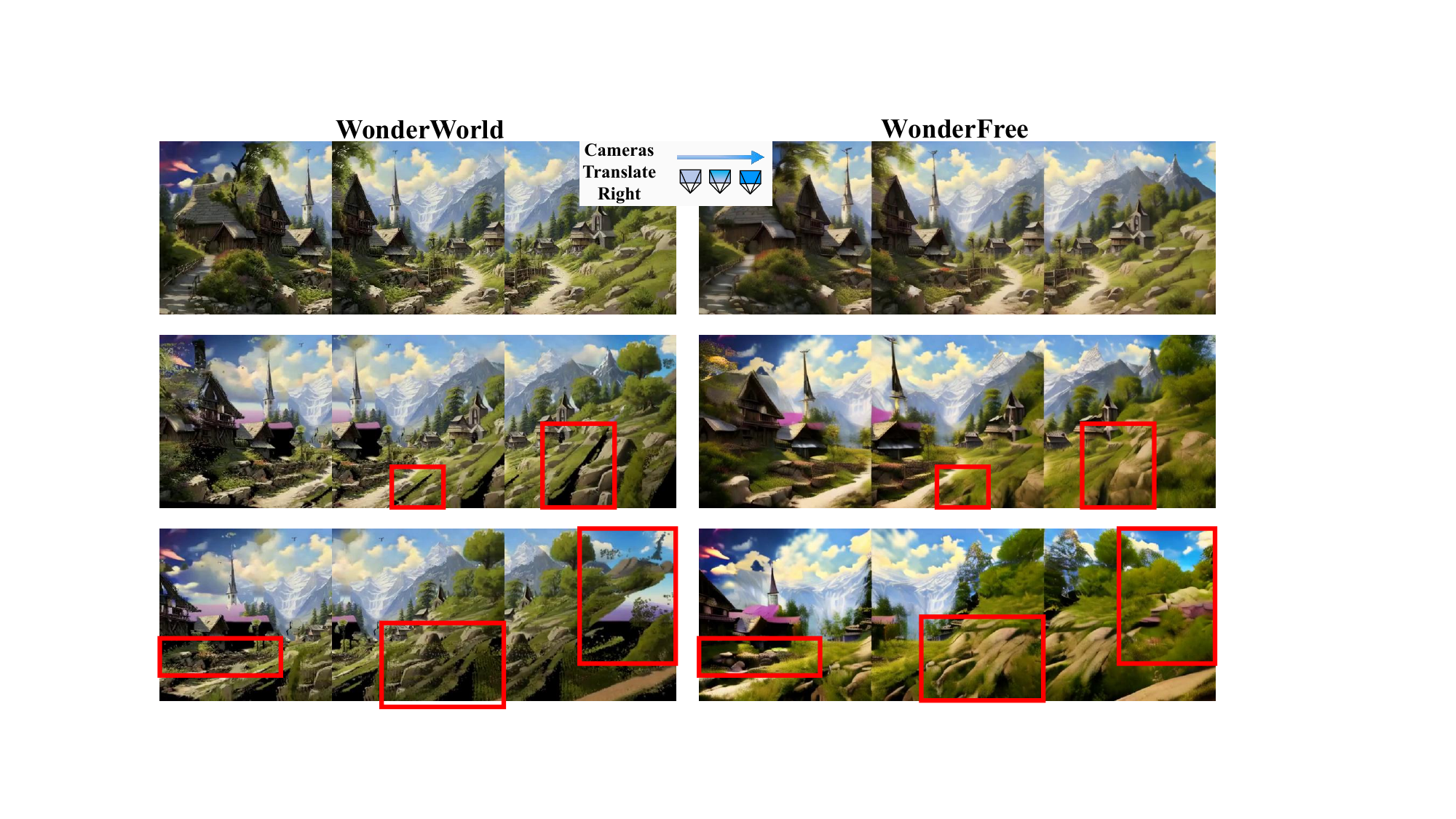}
\includegraphics[width=0.99\textwidth]{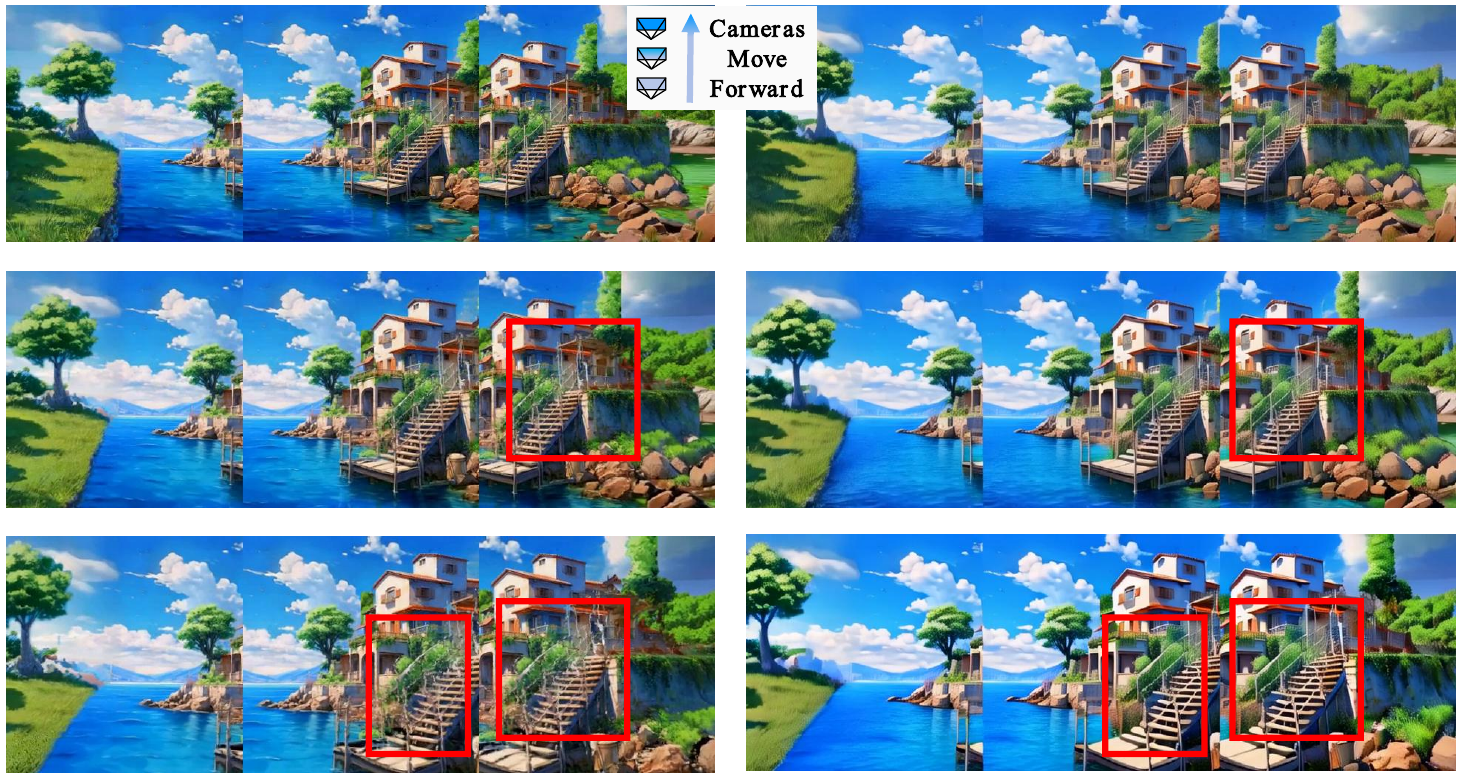}
\includegraphics[width=0.99\textwidth]{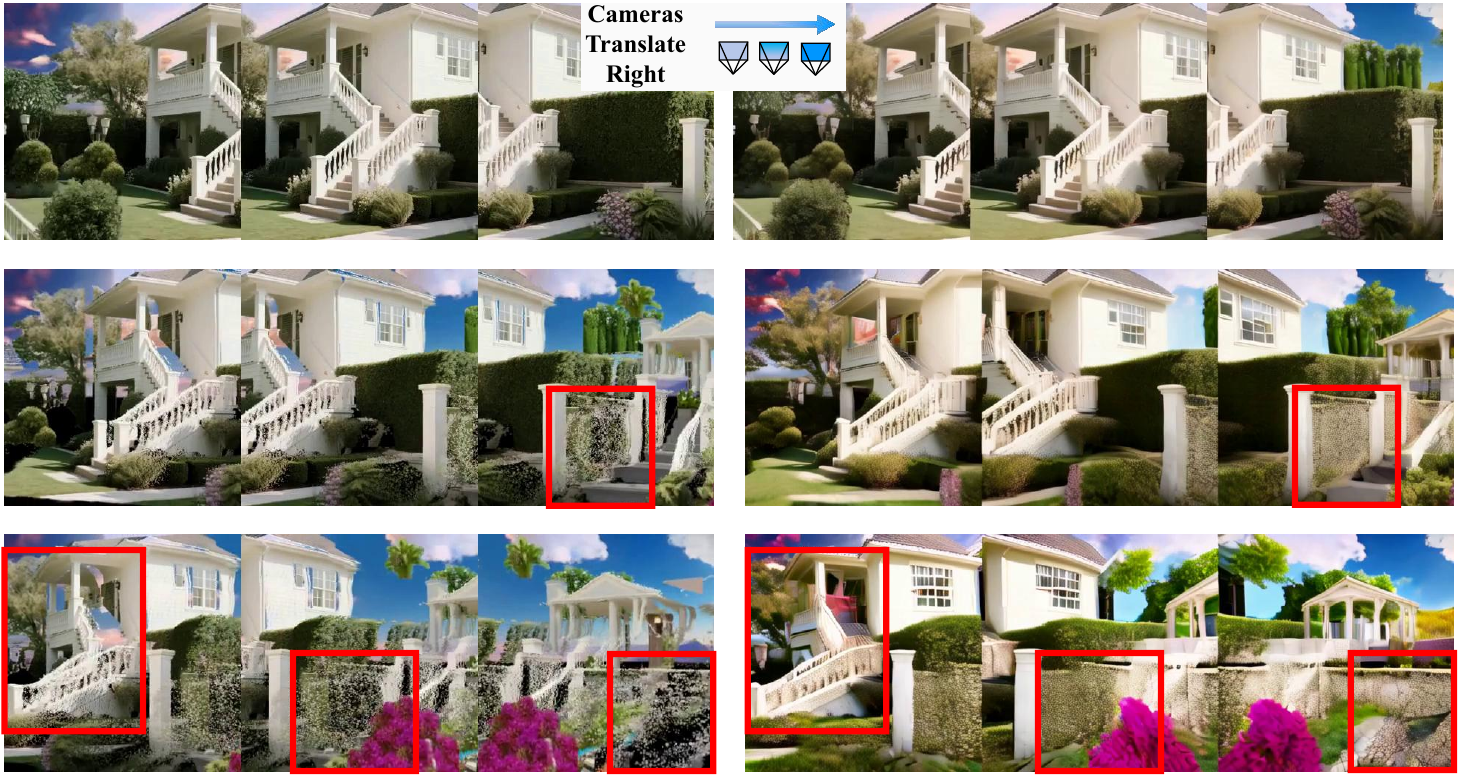}
\caption{Qualitative examples.}
\label{fig:q7} 
\vspace{-1.5em}
\end{figure}

As shown in Fig. \ref{fig:q1} to Fig.~\ref{fig:q7}, we present comparisons between WonderFree and WonderWorld~\cite{wonderworld} under various styles and different trajectories, clearly demonstrating the superiority of WonderFree. Meanwhile, we provide two videos, one comparing WonderWorld~\cite{wonderworld} and WonderFree (\texttt{videos/comparison1.mp4}), and the other illustrating how WonderFree effectively refines the coarse 3D world compared to the original version (\texttt{videos/comparison2.mp4}).

\end{document}